\documentclass{article}

\usepackage[preprint]{neurips_2024}

\usepackage[utf8]{inputenc} 
\usepackage[T1]{fontenc}    
\usepackage{hyperref}       
\usepackage{url}            
\usepackage{graphicx}
\usepackage{booktabs}       
\usepackage{amsfonts}       
\usepackage{nicefrac}       
\usepackage{microtype}      
\usepackage{natbib}
\usepackage{multirow}
\usepackage{enumitem}
\usepackage{authblk}
\usepackage{colortbl} 
\usepackage[table,xcdraw]{xcolor}
\usepackage{enumitem}
\newcommand{\NAME}{\textsc{\texttt{DataS$^3$}}}

\newcommand{\needcite}[1]{\textcolor{red}{CITE}}
\usepackage{amsmath} 
\usepackage{cleveref}
\usepackage{booktabs}
\usepackage{tikz}
 
\usepackage{tabularx}
\DeclareMathOperator*{\argmin}{arg\,min}

\newcommand{\Alg}{\texttt{SubsetSelection-ALG}}
\makeatletter
\renewcommand\AB@affilsepx{ , } 
\makeatother

\def\paperID{9020} 
\def\confName{ICCV}
\def\confYear{2025}

\title{DataS$^3$: Dataset Subset Selection for Specialization}

\author[1]{Neha Hulkund}
\author[5,1]{Alaa Maalouf}
\author[1,2]{Levi Cai}
\author[1,2]{Daniel Yang}
\author[1]{Tsun-Hsuan Wang}
\author[3]{Abigail O'Neil}
\author[1]{Timm Haucke}
\author[3]{Sandeep Mukherjee}
\author[4]{Vikram V. Ramaswamy}
\author[7]{Judy Hanwen Shen}
\author[9,10]{Gabriel Tseng}
\author[11]{Mike Walmsley}
\author[1]{Daniela Rus}
\author[3]{Ken Goldberg}
\author[6]{Hannah Kerner}
\author[3,8]{Irene Y. Chen}
\author[2]{Yogesh Girdhar}
\author[1]{Sara Beery}

\affil[1]{MIT} 
\affil[2]{Woods Hole Oceanographic Institution}
\affil[3]{UC Berkeley}
\affil[4]{Princeton University}
\affil[5]{University of Haifa}
\affil[6]{Arizona State University}
\affil[7]{Stanford University}
\affil[8]{UCSF}
\affil[9]{McGill University}
\affil[10]{Mila - Quebec AI Institute}
\affil[11]{University of Toronto}

\begin{document}

\newcommand{\TS}{\textsc{DS3}}
\newcommand{\problem}{scientific problem}
\newcommand{\domain}{deployment}
\newcommand{\numberofsets}{five}

\maketitle
\vspace{-20pt}
\begin{figure*}[h]
\begin{center}
\includegraphics[width=0.965\linewidth]{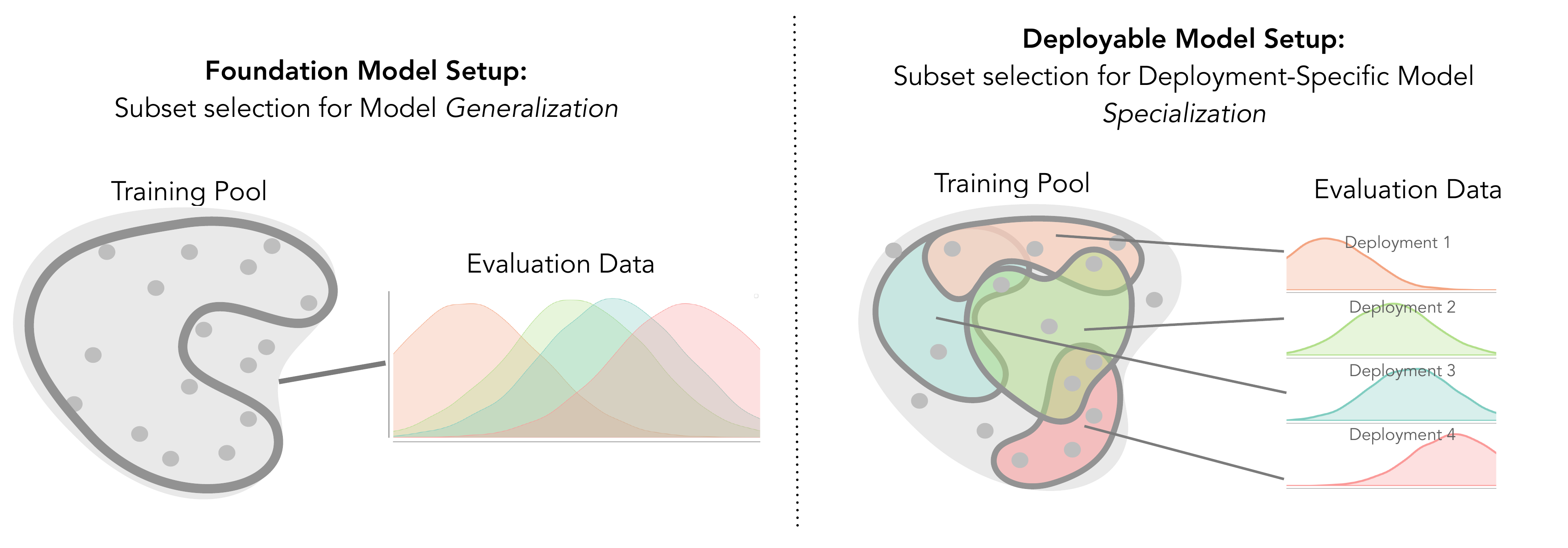}
\caption{Foundation model training aims for broad generalization, by using all data available, usually from massive internet-scale datasets.  In practice, we find these models are often suboptimal for specific deployments, which may exhibit different distributions over categories or data characteristics from the general training data pool. Dataset subset selection for specialization seeks to identify model training subsets closely aligned with the target deployment, achieving superior performance under the given distribution and attribute shifts.
}
    \label{fig:teaser}
\end{center}
\end{figure*}

\begin{abstract}
\vspace{-5pt}
In many real-world machine learning (ML) applications (e.g. detecting broken bones in x-rays or species in camera traps), models need to perform well on specific \domain{}s (e.g. a specific hospital or national park) rather than the domain broadly. However, deployments often have imbalanced, unique data distributions. Discrepancies between training and deployment distributions lead to suboptimal performance, highlighting the need to curate training data for specialization. We formalize \textbf{dataset subset selection for specialization (\TS)}: given a training set drawn from a general distribution and a (potentially unlabeled) query set drawn from the desired \domain{}-specific distribution, the goal is to select a subset of the training data that optimizes \domain{} performance. 

We introduce \NAME{}; the first dataset and benchmark designed specifically for the \TS{} problem. \NAME{} encompasses \numberofsets{} \textit{real-world} application domains, each with a set of distinct \domain{}s to specialize in.   
We conduct a comprehensive study evaluating algorithms from various families—including coresets, data filtering, and data curation, and find that methods trained on general distributions consistently fail to specialize and perform optimally on \domain{} tasks. Additionally, we demonstrate the existence of curated (deployment-specific) subsets that outperform training on all available data by up to $51.3\%$.
Our benchmark highlights the critical role of tailored dataset curation in enhancing performance and training efficiency on \domain{}-specific distributions, which we posit will only become more important as global, public datasets become available across domains and ML models are deployed in the real world.
\end{abstract}

\vspace{-6mm}
\section{Background and Motivation}
Machine learning models are typically trained on large datasets with the assumption that the training distribution closely matches the distribution of the \domain{} where the model will be applied. However, in real-world applications, \domain{} data distributions often diverge from general and/or global 
training set distributions~\cite{shen2024dataadditiondilemma,Taori2020MeasuringRT}. 
Selecting relevant data subsets aligned with specific \domain{}s is crucial to maximize field performance. The problem of \textit{data subset selection for specialization} (\TS{}) is thus critical: given all available training data for a domain and a small (usually unlabeled) query set that represents the desired \domain{}, the goal is to identify a subset of the training data, such that training the ML model on this subset maximizes performance on the \domain{} distribution.

\noindent\textbf{Real world example.} Consider a wildlife ecologist who aims to build a classifier to detect the presence of invasive rodents in camera trap images collected at the Channel Islands. Existing labeled training data on invasive rodents in this context is limited, as they have been mostly eradicated by previous successful conservation action, thus training a classifier from scratch is likely to be unsuccessful. A common approach is to finetune a general pre-trained model (such as ViT or CLIP) on all \textit{relevant} camera trap data. But \textit{what does "relevant data" mean?} Would using similar species data from other camera trap locations (perhaps on the mainland) improve performance, or introduce noise? What about including data from non-similar species at that location? While adding data to a training set can sometimes improve performance, it can also decrease individual subgroup performance in a biased way~\cite{Compton2023WhenMI} and introduce spurious correlations that can enable models to learn potentially dangerous ``shortcuts,'' resulting in biased predictions, shown across various domains~\cite{Geirhos2020ShortcutLI, Badgeley2018DeepLP, Wang2021GeneralizingTU, Beery2022TheAA}.

\noindent\textbf{The Gap: General datasets vs. \domain-specific needs.}
To address the gap between general models and specific deployment needs, we highlight the need for research on \TS{}: the development of methods that select optimal training data for deployment-specific model specialization. Similarly, subset selection methods are evaluated on standard CIFAR10/100~\cite{krizhevsky2009learning} and ImageNet~\cite{imgnet} datasets, where test and validation sets have similar distribution to their training sets. Current benchmarks for data filtering \cite{gadre2024datacomp} focus on generalization across many tasks, in contrast to specialization for a particular deployment. While these works are valuable, they do not capture or enable progress on the  \TS{} challenge. 

\noindent \textbf{Our contributions.} We propose \NAME{}, a comprehensive benchmark to evaluate and compare \domain{}-specialization subset selection methods. Our key contributions are the following:
\newcommand{\numberofbaseline}{1500}
\begin{enumerate}[label=(\roman*)]
    \item \NAME: A benchmark of \numberofsets{} datasets for evaluating algorithms on the \TS{} problem. The datasets represent real-world scientific and engineering applications from different fields. Each dataset includes multiple realistic \domain{}s, e.g., the camera traps dataset aims to categorize species, and the \domain{}s to specialize to are geographic locations where scientists want to analyze species, e.g., Central Africa or Southeast Asia.
    \item We show that selecting a well-curated subset can consistently outperform models trained on the entire dataset with proof-of-concept manually curated subsets of the training data for each \domain{}.
    \item An extensive experimental study comparing current SOTA subset selection methods on \NAME. 
    After training a suite of baselines, our results clearly show that current subset selection methods fail on \TS{}, highlighting the need for our \NAME{} benchmark.
    
\end{enumerate}
\section{Problem Statement}
\noindent\textbf{\TS{} problem formulation. }Let $X$ be a ground set of data points, $T \subset X$ be a given \textit{training set} drawn from a {training (pool) distribution} $P_T$ over $X$, and let $Q \subset X$ be a \textit{query set} drawn from the desired \textbf{\domain{}-specific distribution} $P_Q$ over $X$. Given a model $\theta$, the objective of \textbf{dataset subset selection for specialization (\TS{})}, is to design an algorithm $\Alg$, which takes $T$ (the training set) and $Q$ (the \domain{} representative query set) as input, and outputs a subset $S^* \subset T$ that minimizes the expected loss of $\theta$ trained on $S^*$ over the desired \domain{}-specific distribution $P_Q$. More formally:

\begin{equation}
    S^* = \argmin_{S \subset T} \mathbb{E}_{q \sim P_Q} \left[ \mathcal{L}(\theta(S), q) \right],
\end{equation}

where $\theta(S)$ denotes the model trained on the subset $S \subset T$, and $\mathcal{L}(\theta(S), q)$ is the loss function evaluated on a single point $q$ sampled from $P_Q$ and the trained model $\theta(S)$. The term $\mathbb{E}_{q \sim P_Q}$ denotes the expected value over the distribution $P_Q$.
Hence, the algorithm $\Alg$ outputs $S^*$, the subset of $T$ that minimizes the expected loss of the entire desired \domain{} distribution $P_Q$. Notably, $\Alg$ can only access the desired deployment-specific distribution via the query set $Q$.

\noindent\textbf{Is the query set annotated/labeled?} This formalization can be divided into two cases: in the first, the query set $Q$ is annotated with a set of labels, formally, $Q$ is a set of $m>0$ pairs $Q = \{(q_1, y_1), \cdots, (q_m, y_m)\}$, where for every $i\in [m]$, $q_i$ is the $i$th feature vector describing the $i$th input, and $y_i$ is it corresponding label/annotation. 
In this case the algorithm $\Alg$ has access to the set of labels $\{y_1,\cdots,y_n\}$. 
In the second scenario, no labels are provided for $Q$, meaning that the $\Alg$ does not have access to the set $\{y_1,\cdots,y_n\}$ and consequently $Q = \{q_1,\cdots,q_m\}$. Annotating $Q$ for any specific deployment requires time, money, and expertise. Thus, \TS{} progress without labels has a high potential for impact.


\noindent\textbf{Should \Alg{} be sample efficient?} The goal of our benchmark is to specialize on a desired deployment distribution. Unlike standard subset selection, where subset size is often a primary concern, our focus is on selecting subsets that yield models best suited for deployment. That said, a smaller subset offers many advantages, such as training efficiency, lower memory/storage, etc, thus, we do analyse these trends as well.

\section{Related Work} \label{sec:related_work}

Traditional data subset selection approaches can be split into two main categories: 1) Data filtering or cleaning, which focuses on refining the dataset to enhance its quality~\cite{zhang2022opt,raffel2020exploring}, and 2) Coresets for dataset subset selection, aimed at reducing training time by a computing a subset that effectively represents the larger training dataset~\cite{killamsetty2021glister,tukan2023provable}.

\noindent\textbf{Data filtering for better learning. } 
Data pruning is widely used in NLP to clean noisy datasets~\cite{anonymous2023when}, often employing filtering and heuristics~\citep{bane-etal-2022-comparison}. Common methods include excluding texts with blocklisted words~\citep{raffel2020exploring}, removing duplicates~\citep{zhang2022opt}, filtering out non-English texts~\citep{raffel2020exploring, rae2022scaling}, and discarding short sentences~\citep{raffel2020exploring, rae2022scaling}. Perplexity-based filtering removes high-perplexity samples considered unnatural and detrimental to performance~\citep{muennighoff2023scaling, wenzek-etal-2020-ccnet, laurençon2023bigscience}.
Although simple filtering can enhance language models~\citep{penedo2023refinedweb, raffel2020exploring}, their effectiveness varies, and some studies report no benefits~\citep{black-etal-2022-gpt, biderman2023pythia}, possibly due to their simplicity. \cite{zhou2024lima} showed that manually selecting a small subset satisfying quality and diversity improves alignment performance.
\noindent\textbf{For vision tasks,} a smaller number of methods have been suggested for data filtering~\citep{sorscher2023neural} to obtain better trainable subsets~\citep{siddiqui2022metadata} through the use of model signals~\cite{mindermann2022prioritized}. 



\noindent\textbf{Coresets for efficient learning.} Subset selection (hitherto referred to as coresets) is common for vision tasks. The goal is to compute a small subset from the training dataset, that approximates training on the full dataset, thus boosting the training process~\cite{braverman2016new,maalouf2022unified}.
Coresets proved to be useful in many applications such as regression~\cite{DasguptaDHKM08,Chhaya0S20,TolochinskyJF22,MeyerMMWZ22,maalouf2019fast}, clustering~\cite{Har-PeledM04,Chen09,HuangV20,jubran2020sets,cohen2022improved}, low-rank approximation~\cite{CohenMM17,BravermanDMMUWZ20,maalouf2021coresets}, support vector machines (SVMs)~\cite{Clarkson10,TukanBFR21,maalouf2022unified}, and for compressing neural networks~\cite{BaykalLGFR22,liebenwein2019provable,Tukan2022provable}. For boosting the training of neural networks,~\cite{coleman2019selection} used proxy functions
to select subsets of training data approximating the training process. Later~\cite{mirzasoleiman2020coresets,MirzasoleimanCL20} developed algorithms to estimate the full gradient of the deep neural network on the training data and then further refined by \cite{killamsetty2021grad,killamsetty2021glister,paul2021diet,wang2020optimizing}. Other methods require a neural network forward pass to get embeddings~\cite {sener2018active,sorscher2022beyond,killamsetty2021retrieve}. All these methods assume the training data well represents the test (\domain{}) data, as the case in known diverse, high-quality vision benchmarks (CIFAR10 and ImageNet). Thus, the aim was to approximate the training data via a subset (coresets) or enhance training (filtering) assuming that the training and testing sets share the same distribution.

\noindent\textbf{Active learning.} There is a rich area of online active learning literature, which continually filters data while training~\citep{ein-dor-etal-2020-active, wang2022unsupervised, yuan-etal-2020-cold, tamkin2022active}, requiring to query an annotator for more labeled data and  oftentimes, rely on properties of the models in-training to select data. Here, we are interested in exploring data subselection prior to training and without knowledge of model weights. 

\noindent\textbf{Benchmarks.} The works most related to ours are~\cite{gadre2024datacomp}, \cite{mazumder2023dataperfbenchmarksdatacentricai} and~\cite{feuer2024select}. DataComp~\cite{gadre2024datacomp} introduces a benchmark where the main challenge is to select the optimal data subset for pretraining \textit{generalization}. It evaluates various data curation strategies using standardized CLIP training code, followed by zero-shot assessments on 38 downstream datasets. \cite{mazumder2023dataperfbenchmarksdatacentricai} has multiple benchmarks across domain-specific data sources, but is again aimed for generalization rather than specialization. \cite{feuer2024select} focuses on image-only models, which are smaller and easier to train to high accuracy. 

\noindent\textbf{Our benchmark. }In contrast to these benchmarks, \NAME{} is specifically designed to evaluate subset selection methods for \textit{\domain{}-specific specialization}, rather than generalization, where the training and testing (\domain{}) data exhibit distributional shifts.

\section{The \NAME{} Benchmark}\label{sec:benchmark}


\begin{figure*}
    \includegraphics[width=\textwidth]{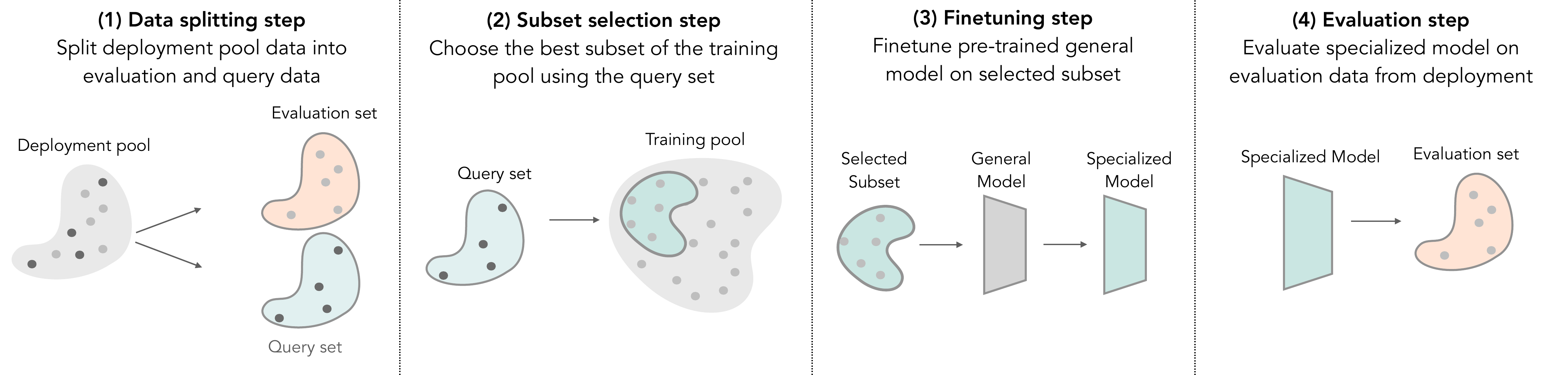}
    \caption{\NAME benchmark process, involving dataset splitting, subset selection, model specialization/finetuning, and then evaluation.}
    \label{fig:benchmark_process}
\end{figure*}


\begin{figure*}
    \centering
    \includegraphics[width=1.0\textwidth]{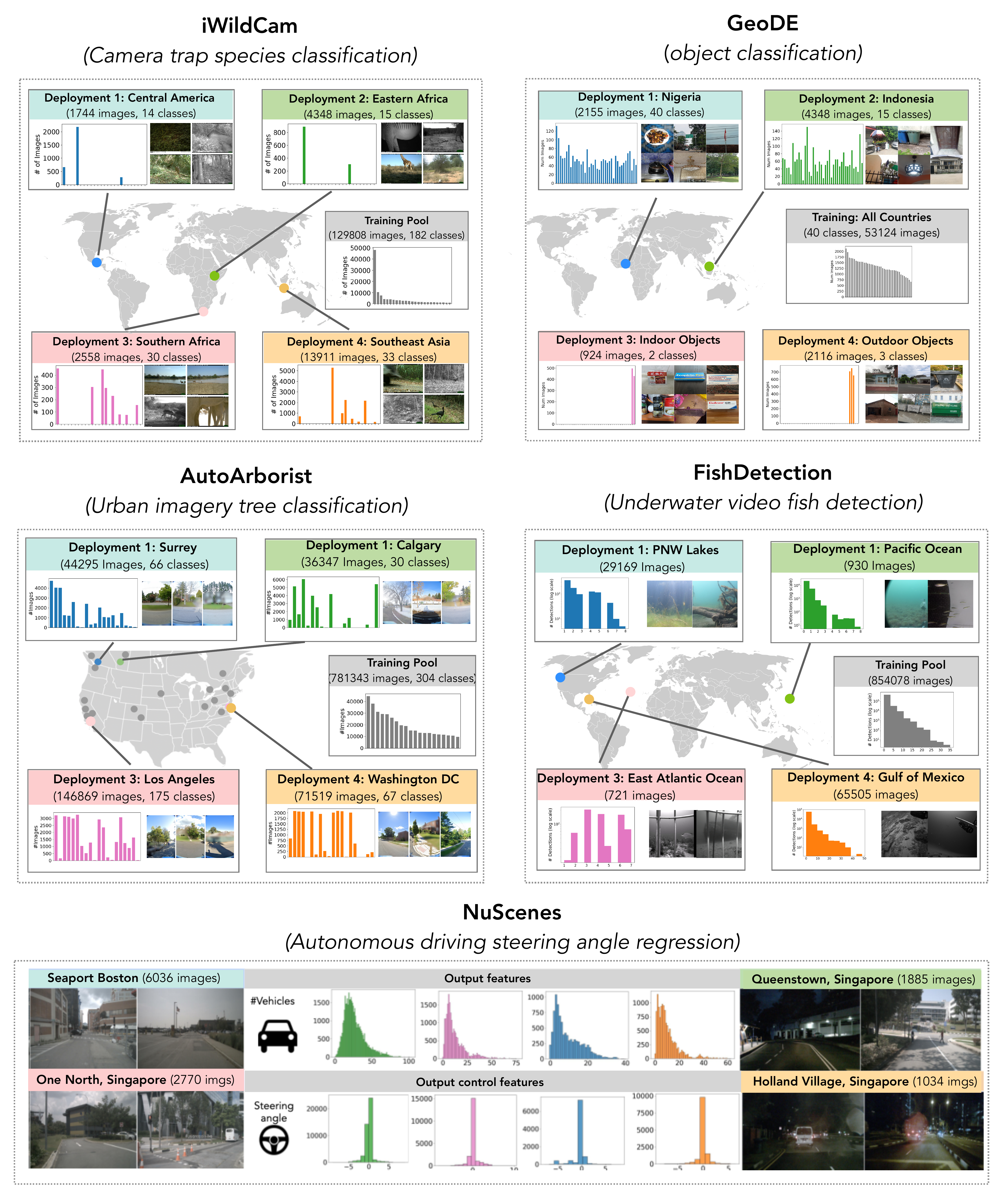}
    \caption{The five datasets in our benchmark: iWildCam, GeoDE, AutoArborist, FishDetection, and NuScenes each have real-world applications in deployment. In iWildCam, GeoDE, and AutoArborist, we show the class distributions of each deployment; in FishDetection, the number of detections per image is shown, and in NuScenes environment/output features. These diagrams show that each dataset has unique challenges in the deployments that lead to a need for model specialization, including long-tailedness (AutoArborist, iWildCam), covariate shift (all),  subpopulation shifts (GeoDE, FishDetection), and more. These axes of variation are described in depth in \cref{sec:benchmark} and further in Apdx \ref{appdx:additional_dataset_details}.}
    \label{fig:dataset_figure}
\end{figure*}
\noindent\textbf{Datasets. } Our benchmark includes \numberofsets{} datasets, each capturing a unique and diverse application of ML: Auto Arborist for tree classification \cite{beery2022auto},  iWildCam for camera trap species identification \cite{beery2021iwildcam}, GeoDE for diverse object classification \cite{ramaswamy2023geode}, NuScenes for autonomous driving footage steering regression ~\cite{caesar2020nuscenes}, and FishDetection for underwater video fish detection ~\cite{viame}. 
Each of these datasets inherently represents many of the real-world challenges that make dataset subset selection a \domain{}-specific problem, including covariate shifts, subpopulation shifts, and long-tailed distributions. 
For each dataset, we provide a proof-of-concept "oracle"~/~knowledge-driven subset that demonstrates the usefulness of dataset subset selection, with improvement over using all the training data. These subsets were created using information that benchmark users are not provided (e.g. metadata, GPS location, region, etc).  Additional details about each dataset can be found in Apdx. \ref{appdx:additional_dataset_details}. 

\subsection{iWildCam}
\textbf{Motivation:} 
Animal populations have declined by $68\%$ on average since 1970~\cite{Staub_2020}. To monitor this biodiversity loss, ecologists deploy camera traps—motion-activated cameras placed in the wild~\cite{camera_traps_best_practice}—and process the data with machine learning models~\cite{norouzzadeh2019deepactivelearningspecies,beery2019efficientpipelinecameratrap}. However, variations in illumination, camera angle, background, vegetation, color, and animal frequencies across different locations cause these models to generalize poorly to new \domain{}s. To specialize models for specific locations, selecting appropriate data subsets for \domain{}-specific (in this case location) specialization becomes essential.

\noindent\textbf{Problem Setting \& Data:} To study this problem, we use the iWildCam 2020 dataset, comprising of $203,029$ images from $323$ different camera traps spread across multiple countries in different parts of the world. The task is multi-class species classification. Concretely, the input \textit{x} is a photo taken by a camera trap, the label \textit{y} is one of $182$ different animal species, and the \domain{} $d$ is an integer that identifies the camera trap that took the photo. These images tend to be taken in short bursts following the motion-activation of a camera trap, so the images can be additionally grouped into sequences of images from the same burst, though our baseline models do not exploit this information, and our evaluation metric treats each image individually. However, a grouped sequence is in the same split of the data (train, test, query) in order to avoid model memorization. Each image is associated with the following metadata: camera trap ID, sequence ID, and datetime. Performance is measured by overall classification accuracy for species identification. The original camera trap data comes from the Wildlife Conservation Society (\href{http://lila.science/datasets/wcscameratraps}{link}).



\noindent\textbf{Deployments:} Our \domain{}s were defined to be split across camera trap locations to simulate the common scenario of researchers setting up new cameras within a region, with poor model generalization on the new cameras \cite{camera_traps_best_practice}. Our train/test split was done randomly across the $200$ locations, with the five downstream test tasks created by clustering by the latitude and longitude of camera GPS location in 4 deployments: (1) Central America, (2) Eastern Africa, (3) Southern Africa, and (4) Southeast Asia. Similar to most other camera trap datasets, iWildCam has significant long-tailed label distributions, with variation in species and backgrounds between locations, as can be seen in Figure \ref{fig:dataset_figure}.

\noindent\textbf{Knowledge-driven Subset:} These subsets were created by only choosing training data from camera locations that are within 100km of the camera locations in the deployments (the relevant geographical area) and eliminating irrelevant classes that are not present in the \domain{}.

\subsection{GeoDE}
\noindent\textbf{Motivation:} Object classification datasets are often constructed by scraping images from the web but contain geographical biases~\cite{Shankar2017NoCW, imagenet,Russakovsky2014ImageNetLS,Kuznetsova2018TheOI,Asano2020NewCD,Brown2020LanguageMA}. Instead of scraping images from the web, GeoDE~\cite{ramaswamy2023geode} crowdsources a dataset that is roughly balanced across 40 different objects and six world regions, showing that common objects (stoves, bicycles, etc), vary in appearance across the world. Crowdsourcing a dataset allows for tighter control over the data distribution. For example, it allows us to target specific regions and objects that are underrepresented within webscraped datasets. However, it can also be prohibitively expensive, limiting the size of such datasets. Thus, it becomes paramount to understand which objects and regions should be targeted within crowdsourced data collection.

\noindent\textbf{Problem setting \& Data:} GeoDE is a diverse dataset of 61,490 images comprising 40 different objects collected from 6 world regions (Africa, Americas, East Asia, Europe, Southeast Asia, West Asia). The associated task is multiclass classification, where the goal is to predict the object depicted in each image.   

\noindent\textbf{{Deployments:}} We propose 4 different \domain{}s: (1) objects in Indonesia, (2) objects in Nigeria, (3) indoor objects, and (4) outdoor objects, as shown in Figure \ref{fig:dataset_figure}. Nigeria and Indonesia were selected as the two countries with the poorest performance, and the indoor/outdoor deployment tasks were selected for enabling model specialization. The training dataset includes images from all countries, and the test data contains only images from Nigeria and Indonesia. 

\noindent\textbf{Knowledge-driven Subset:} These subsets were generated by selecting data from the relevant countries/categories in the training data, ie. only selecting African subcontient data for the Nigeria deployment, Asian subcontinent data for the Indonesia deployment, and  indoor/outdoor objects within the training pool for these deployments.

\subsection{Auto Arborist}
\label{sec:autoarborist}
\textbf{Motivation:} 
Ecological imagery for environmental monitoring and Earth observation provides policymakers with critical, data-driven insights to support climate adaptation~\cite{Brandt2016AFF}.  Automated tree classification, for instance, offers substantial benefits for humanitarian aid, disaster relief, forestry, agriculture, and urban planning, supporting applications in city planning, resource management, and environmental monitoring. This task is associated with fundamental challenges such as noisy labels, non-iid data, fine-grained and long-tailed class distribution, and geospatial distribution shift. These challenges lead to a need for specialization of models where general-purpose models fail. 

\noindent\textbf{Problem Setting \& Data:} The Auto Arborist dataset is a multi-view, fine-grained visual tree categorization dataset containing street-level images of over 1 million public zone trees from $300$ genus-level categories across $23$ major cities in the US and Canada.  

\noindent\textbf{Deployments:} Deployments in Auto Arborist correspond to the development models for use by individual cities. The deployment cities of (1) Surrey with 66 distinct tree genus classes, (2) Calgary with 30 classes, (3) Los Angeles with 175 classes, and (4) Washington DC with 67 classes were chosen due to their diverse climates, species distributions, and urban structures, as seen in Figure \ref{fig:dataset_figure}. Surrey and Calgary were treated as our in-distribution (ID) deployments, with some of these cities data in the training pool. Washington DC and LA were the out-of-distribution deployments, with no city data in the training pool. 


\noindent\textbf{Knowledge-driven Subset:}  We used the relevant data from Surrey and Calgary in the training pool for these ID deployments. Accordingly, we used data from San Francisco and San Jose for Los Angeles and Charlottesville, Pittsburgh, and New York for Washington DC. Label distribution shift   and covariate shift are visualized in Figure \ref{fig:autoarborist_deployments} and~\ref {fig:autoarborist_viz}, respectively.

\subsection{Fish Detection}
\textbf{Motivation:} Climate change, pollution, and overfishing continue to threaten marine biodiversity and fishery stocks across the globe \cite{UN_BNJJ, di2022sustainable}. Marine imagery, collected from baited remote underwater videos (BRUV) or remotely-operated and autonomous underwater vehicles (ROVs and AUVs), is an increasingly common resource utilized by marine scientists to monitor and assess fish stocks and biodiversity. Machine-learning methods provide faster ways to perform analysis; however, are difficult to apply across environmental settings due to differences in lighting, turbidity, species, vegetation, camera sensors, etc. \cite{borremans2024report, jerlov1976marine, akkaynak2019sea}.

\noindent\textbf{Problem Setting \& Data:}  We use the public VIAME FishTrack23 dataset \cite{viame} consisting of 854,078 images across various environmental settings, ranging from freshwater rivers to deeper benthos. Specifically, the task is to predict bounding box localizations around every fish present in each image. Performance is measured by mAP across various IoU thresholds. Most of the images across all datasets are taken from video streams, and can be grouped as such, that were deployed primarily on camera traps, both baited and unbaited.

\noindent \textbf{Deployments:} Deployments are split according to the subsets of the VIAME dataset, which roughly correspond to geographic regions. Train, test and subset splits are either taken as provided or randomly sampled frames from each dataset, roughly corresponding to: (1) freshwater Pacific Northwest lakes; (2) Pacific Ocean; (3) East Atlantic Ocean; and (4) Gulf of Mexico.

\noindent \textbf{Knowledge-driven Subset:} For each deployment, we use the subset in the relevant geographical area (e.g., images from Gulf of Mexico for the Gulf of Mexico deployment).

\newcommand{\carsetname}{NuScenes}
\subsection{\carsetname}
\noindent\textbf{Motivation:} 
End-to-end autonomous driving systems streamline vehicle control by directly mapping sensory inputs, such as images, to control outputs like steering angles~\cite{wang2024drive}. Adapting these systems to specialize in particular streets or environments is made easier as a single model encompasses the full system. Thus, training this model to specialize in a specific environment brings advantages, capturing detailed local road layouts, typical traffic patterns, area-specific obstacles, and more. 


\noindent\textbf{Problem Setting \& Data:} We explore vision-based control for self-driving across diverse environments (e.g., different city areas) and driving scenarios (e.g., pedestrians crossing, construction zones), formulated as a regression task. This dataset includes $88,461$ images from the NuScenes dataset, subsampled from the image sweeps at a rate of 2. The images were captured from a video stream recorded while driving a car. Each image is paired with a steering angle control from the CAN bus, synchronized with the sensor timestamps of both the camera and CAN bus data. To label each image with the correct steering angle, we apply 1D interpolation to create a continuous function of the steering angle and query it based on the camera’s timestamp. The steering angle, measured in radians, ranges from -7.7 to 6.3, with 0 indicating straight driving, positive values indicating left turns, and negative values indicating right turns. To ensure alignment between images and steering control data, samples with vehicle velocities below 1 m/s are removed. The model's goal is to predict a single scalar value representing the car's steering angle. Performance is evaluated in an open-loop manner using metrics like mean squared error.

\noindent\textbf{Deployments:} Deployments are organized by the geographic locations where the data was collected, including (1) Boston Seaport, (2) Singapore Holland Village, (3) Singapore One-North, and (4) Singapore Queenstown. While all tasks are based on expert demonstrations of driving and general driving behaviors, each location presents varying environmental features—such as vegetation, road types, roadside infrastructure, and weather—as well as differences in driving style and road regulations. Train/test splits are randomly sampled within each \domain{}.

\noindent\textbf{Knowledge-driven subset: }
Since this training pool is a combination of the four deployment locations, we simply use the relevant location's data as the training subset. For example, we use the subset of the training pool with Boston Seaport data for the Boston Seaport deployment.


\subsection{Benchmark Pipeline}
\vspace{-1mm}
Within our benchmark, each dataset has a two-step process for evaluation:
\begin{enumerate}[label=(\roman*)]
    \item Given a small query set representing the \domain{} data (we consider both labeled and unlabeled query sets), curate a subset of data from the training for a specific \domain{}. 
    \item Finetune/train a fixed model on the chosen subset from the training pool and evaluate on the \domain{} (test) set.
\end{enumerate}

For each dataset, we fix the training procedure for all subsets, fixing model architecture, optimizers, and loss functions. We run a small hyperparameter sweep for each training subset across batch sizes $\{32,64,128\}$ and learning rates $\{0.01,0.001,0.0001\}$ for each deployment. For all classification/regression datasets, we use ResNet50 for full-finetuning \cite{He2015DeepRL} (\cref{tab:main_results}) and a ViT for LoRA finetuning (Apdx Table \ref{tab:lora_results}), as well as a ViT \cite{Dosovitskiy2020AnII} for linear probes (Apdx Table \ref{tab:linear_probe_results}). For the detection dataset, we use a YOLOv8n model, using default parameters, though we subsample images to 640p. Full details are in Apdx \ref{appdx:additional_training_details}.

\subsection{Metrics} \label{sec:evaluation}
Participants are evaluated across 12 deployments from \numberofsets{} datasets, as outlined in Section \ref{sec:benchmark}. For the classification task datasets of GeoDE, Auto Arborist, and iWildCam, we report accuracy for each deployment, for the regression task dataset NuScenes, we report mean squared error, and for the detection task FishDetection, we report mAP50. For each deployment, we evaluate participants of the benchmark on overall accuracy of training subset; we also report subset size -- while the less data used the better, we mainly focus on optimal performance.

\section{Baselines}
We compare performance of coreset/data filtering algorithms for dataset subset selection across our benchmark, across different scenarios: (a) access to an unlabeled query set, and (b) access to a labeled query set. We also curate a third category, (c), which leverages domain expertise to generate expert-selected subsets, in order to demonstrate the existence of better-than-all subsets for these deployments.

\vspace{2pt}
\noindent
\textbf{Non-subset comparisons:}\\
\noindent\textit{\underline{No filtering:}} Performance of a model trained on the entire training pool, without any filtering. 

\noindent\textit{\underline{Query Sets:}} As a comparison, we also include performance of a model trained directly on the labeled query set for each deployment. Note that this would require access to query labels, which are not always available. When labels are available, performance of models trained on the small query sets are often poor, hence the value of learning from larger-scale general-pool data. As a logistical point, none of the baselines we show in our results train on query set data.

\vspace{2pt}
\noindent\textbf{Expert-Driven Subsets:} We contribute curated, "expert knowledge" subsets using domain knowledge and/or metadata. We find these knowledge-guided subsets often outperform using all samples in the training pool (no filtering). The creation of these subsets is described per-dataset in~\cref{sec:benchmark}.


\vspace{2pt}
\noindent\textbf{Unlabeled-query baselines:}\\
\noindent\textit{\underline{Image-alignment (Image-Align):}} We take the cosine similarity between the training and query embedding space, using examples that exceed a threshold for at least $x$ samples, where $x$ is a hyperparameter chosen from \{1,10,100\}.

\noindent\textit{\underline{Nearest neighbors features (Near-Nbors):}} To better align our method with the downstream deployment, we explore using examples whose embedding space overlaps with the query set of data. To do so, we cluster image embeddings extracted by an OpenAI ViT model for each image into $1000$ clusters using Faiss \cite{faiss_gpu}. Then, we find the nearest neighbor clusters for every query set example and keep the training cluster closest to each query set cluster. This method was inspired by the similar DataComp baseline \cite{gadre2024datacomp}.

\vspace{2pt}
\noindent\textbf{Labeled-query baselines:}\\
\noindent\textit{\underline{CLIP score filtering (CLIP-score):}} We also experiment with CLIP score filtering, using examples that exceed a threshold for cosine similarity between CLIP image and text similarity. Text for each image was created with manual captioning (e.g. for iWildCam, \textit{"This is a camera trap image of a lion taken at time 10-2-2016 at 04:26:13 in Nigeria"}). We select the subset that exceeds a threshold of CLIP-score similarity, with the threshold calculated for subsets that make up $25\%, 50\%$, $75\%$, and $90\%$ of the dataset.

\noindent\textit{\underline{Matching relative frequency (Match-Dist): }} We explore having access to the relative frequency of each label in the downstream deployment. For example, a domain expert at a national park might know the relative frequency of species (deployment-specific domain knowledge). We create subsets by sampling $25\%$, $50\%$, $75\%$, and $90\%$ of the training pool to match the label distribution of the deployment. 

\noindent\textit{\underline{Matching labels (Match-Label):}} Similarly, a domain expert may know the classes present in the downstream deployment. For example, a domain expert at a national park might know the species present (deployment-specific domain knowledge) that we can utilize for dataset subset selection. For these subsets, we simply remove the classes present in the training pool that are not present in the testing pool. 



\section{Results and discussion}

\begin{table*}[]
\resizebox{\linewidth}{!}{
\begin{tabular}{@{}c|c|c|cc|c|cc|ccc}
\specialrule{.2em}{.1em}{.1em} 
\multicolumn{1}{c|}{\multirow{2}{*}{Dataset}} & \multicolumn{1}{c|}{\multirow{2}{*}{Metric}} & \multicolumn{1}{l|}{\multirow{2}{*}{Deploy \#}} & \multicolumn{2}{c|}{Non subset} & 
\multicolumn{1}{c|}{\multirow{2}{*}{Knowledge-driven}} & \multicolumn{2}{c|}{Unlabeled query set} & 
\multicolumn{3}{c}{Labeled query set} 
\\ \cline{4-5} \cline{7-11}
\multicolumn{1}{l|}{} & 
\multicolumn{1}{l|}{} & 
\multicolumn{1}{l|}{} &
Query-set & All-data &  & Image-Align & Near-Nbors & CLIP-score & Match-Label & Match-Dist \\
\specialrule{.2em}{.1em}{.1em}
\multirow{4}{*}{GeoDE} & \multirow{4}{*}{Acc (\#)} & Deploy 1 & 0.87 (500) & 0.89 (53k) & \textbf{0.92 (2.9k)} & 0.88 (26k) & 0.88 (48k)& 0.89 (40k)& 0.88 (53k)& 0.89 (48k)\\
 & & Deploy 2 & 0.45 (500) & 0.89 (53k)& \textbf{0.91 (2.6k)} & 0.90 (26k) & 0.89 (48k)& 0.90 (40k)& 0.90 (53k)& 0.88 (27k)\\
 & & Deploy 3 & 0.95 (500) & 0.82 (53k)& \textbf{0.85 (1.4k)} & 0.85 (24k) & 0.76 (48k)& 0.84 (40k)& 0.83 (1.4k)& 0.88 (48k)\\
 & & Deploy 4 & 0.83 (500) & 0.83 (53k)& \textbf{0.85 (2.6k)} & 0.79 (24k) & 0.78 (48k)& 0.83 (40k)& 0.84 (2.6k)& 0.83 (13k)\\
  \hline 
\multirow{4}{*}{iWildCam} & \multirow{4}{*}{Acc (\#)} & Deploy 1 & 0.70 (301) & 0.66 (130k) & 0.65 (8.5k) & 0.56 (36k) & 0.50 (117k)& 0.50 (97k)& 0.74 (8.1k)& \textbf{0.74 (117k)} \\
 & & Deploy 2 & 0.78 (302) & 0.34 (130k)& 0.35 (9.2k) & 0.44 (45k) & 0.47 (98k)& 0.46 (97k)& 0.35 (55k)& \textbf{0.49 (65k)} \\
 & & Deploy 3 & 0.44 (301) & 0.72 (130k)& 0.75 (19k) & 0.54 (24k) & 0.45 (98k)& 0.42 (97k)& 0.72 (60k)& \textbf{0.75 (117k)} \\
 & & Deploy 4 & 0.46 (309) & 0.66 (130k)& 0.67 (21k) & 0.60 (22k) & 0.60 (33k)& 0.29 (97k)& 0.69 (57k)& \textbf{0.74 (33k)} \\
  \hline 
\multirow{4}{*}{AutoArborist} & \multirow{4}{*}{Acc (\#)} & Deploy 1 & 0.16 (1.5k) & 0.35 (781k) & \textbf{0.86 (70k)} & 0.38 (44k) & 0.39 (391k)& 0.38 (47k)& 0.67 (368k)& 0.74 (703k)\\
 & & Deploy 2 & 0.20 (1.5k) & 0.48 (781k)& \textbf{0.86 (123k)} & 0.11 (49k) & 0.14 (703k) & 0.14 (47k)& 0.65 (532k)& 0.56 (391k)\\
 & & Deploy 3 & 0.12(1.5k)  & 0.16 (781k)& \textbf{0.38 (35k)} & 0.16 (46k) & 0.10 (703k)& 0.17 (47k)& 0.16 (534k)& 0.23 (703k)\\
 & & Deploy 4 & 0.12 (1.5k) & 0.14 (781k) & \textbf{0.39 (26k)} & 0.10 (48k) & 0.11 (391k)& 0.11 (47k)& 0.10 (527k)& 0.23 (195k)\\
  \hline 
\multirow{4}{*}{NuScene} & \multirow{4}{*}{MSE (\#)} & Deploy 1 & 0.063 (6.0k) & 0.050 (100k) & \textbf{0.029 (20k)} & 0.040 (35k) & 0.040 (90k)& 0.073 (31k)& - & - \\
 & & Deploy 2 & 0.070 (1.0k) & 0.021 (100k) & 0.049 (4.6k) & 0.15 (17k) & 0.042 (90k) & \textbf{0.032(31k)} & - & - \\
 & & Deploy 3 & 0.089 (2.7k) & 0.068 (100k) & \textbf{0.038 (10k)} & 0.049 (28k)  & 0.13 (90k)& 0.071 (31k)& - & - \\
 & & Deploy 4 & 0.12 (1.9k)& 0.048 (100k)  & \textbf{0.039 (7.0k)} & 0.086 (26k) & 0.39 (90k)& 0.050 (31k)& - & - \\
 \hline
 \multirow{4}{*}{FishDetection} & \multirow{4}{*}{mAP50 (\#)} & Deploy 1 & 0.22 (500) & 0.68 (841k) & \textbf{0.69 (179k)} & 0.50 (630k)& 0.60 (103k)& - & - & - \\
 & & Deploy 2 & 0.26 (600) & 0.32  (841k)  & \textbf{0.45 (152k)} & 0.31 (630k)& 0.40 (120k) & - & - & - \\
 & & Deploy 3 & 0.13 (541) & 0.32  (841k)  & \textbf{0.39 (6.0k)} & 0.28 (630k)& 0.23 (204k)& - & - & - \\
 & & Deploy 4 & 0.079 (519) & 0.59  (841k)  & \textbf{0.60 (320k)} & 0.54 (630k)& 0.39 (45k)& - & - & - \\
\specialrule{.2em}{.1em}{.1em}
\end{tabular}
}
\caption{
Best-performing subsets across hyperparameters for baseline methods across all datasets and deployments (abbreviated as Deploy) for YOLOv8 full-finetuning for FishDetection and ResNet50 full-finetuning for the rest. Accuracy is reported for the classification tasks of GeoDE, iWildCam, and Auto Arborist, mAP50 for FishDetection (greater is better), and MSE for NuScenes (smaller is better). We include subset size in parentheses. We include results for \textbf{ViT LoRA finetuning and ViT linear probes} in Apdx. \ref{appdx:additional_results} in Table \ref{tab:lora_results} and Table \ref{tab:linear_probe_results}, which display similar trends. Match-Dist and Match-Label are not applicable for NuScenes, as it is a regression task and does not have clear classes/labels for these methods. FishDetection only uses the unlabeled query set, as the ground truth is bounding boxes, rather than labels themselves. Baselines are distinguished from one another by their access to information, with each baseline having access to expert knowledge, or a labeled/unlabeled query set. We do not report the random baseline in this table, but demonstrate results in Appendix \ref{appdx:additional_results} as it mainly refers to subset size. Well-chosen subsets outperform using all the training data in each deployment, indicated in bold. }
\label{tab:main_results}
\end{table*}

\noindent\textbf{Well chosen subsets outperform training on all data}. The knowledge-driven subsets in Table~\ref{tab:main_results} show that \domain{}-specific well-chosen subsets of the data can significantly outperform models trained on all the data,  with improvements in deployment accuracy up to $3.6\%$ for GeoDE, $11.9\%$ for iWildCam, $51.3\%$ for Auto Arborist, a $0.03$ reduction in MSE for NuScenes, and $0.13$ increase in mAP50 for FishDetection. Even when the knowledge-driven subsets underperform all training data, as in NuScenes Deployment 2, there exist subsets from other baselines that outperform using all the data. Due to the extreme long-tailed nature and significant label distribution shift between the training pool and deployments of iWildCam and Auto Arborist, well-chosen subsets improve performance significantly. This indicates that using "irrelevent" data from the training pool is actively harmful to performance for specialized deployments, compared to a closer in-distribution subset. As an example, the iWildCam's training pool contains many Thompson's Gazelle, but only Deployment 2 has Thompson's Gazelle present. Accordingly, Deployments 1, 3, and 4 had more improvement between all data and knowledge-driven subsets than Deployment 2 since the former had a greater label distribution shift from the training pool. 

\noindent\textbf{There is a need for unsupervised methods for dataset subset selection for specialization.}
While the knowledge-driven subsets in Table~\ref{tab:main_results} demonstrate that a well-chosen subset \textit{does exist} for all deployments, finding this subset without extra knowledge is still an open problem. Some of our baselines require access to query labels, this requirement can in many cases be unrealistic in the deployable ML setting (labels can be expensive or difficult to collect). The two unsupervised baselines, the nearest neighbors and image alignment methods, do not perform optimally on the deployments, often underperforming using all the training data.  Our benchmark opens up the line of research for potential unsupervised methods for this data subset selection process. 

\noindent\textbf{Training on more data has diminishing returns.}
For all deployments, we see that we can achieve near-optimal performance with subsets of the data. The knowledge-driven subsets are significantly smaller than the total training data size, with the average percentage of the total training pool used being $4\%$ for GeoDE, $11\%$ for iWildCam, $8\%$ for AutoArborist, $10\%$ for NuScenes, and $20\%$ for FishDetection. Appendix \ref{appdx:additional_results} shows that even $25\%$ of the data can perform near-optimally in some cases, with little performance loss with $50\%$ of the data on the algorithmic baselines. Overall, these results demonstrate that greater efficiency for training specialized ML models is possible, potentially reducing computational and data storage burdens in deployable settings. We hypothesize this is because many deployments have significant distribution shift from the training pool, so as the data added gets farther from the deployment distribution, it becomes less relevant for optimal performance. 




\section{Conclusions}
We present \NAME{}, a benchmark to explore model specialization via dataset subset selection for scientific and engineering domains, and provide:  \textbf{(1)} a test suite for the problem across \numberofsets{} ML application domains, each represented by a dataset containing a general training data pool and 4 distinct \domain{} scenarios \textbf{(2)} expert- and knowledge-guided subsets for each deployment which outperform training on all data, sometimes by a significant margin, demonstrating the value of specialized training data curation \textbf{(3)} an extensive experimental study highlighting that current methods for subset selection, designed for generalization instead of specialization, do not perform well on \NAME{}. 

We find that there does not currently exist a winning method that performs well across multiple domains/datasets, posing an open challenge to the research community. While well-performing subsets exist via expert knowledge, models without access to labeled query sets systematically underperform. We also find that certain datasets are more challenging than others-- perhaps different subset selection methods are necessary for different domains or types of shifts.

\section{Future Work}
We aim to open this benchmark as  a testbed for algorithmic development for dataset subset selection for domain-specific needs, focusing particularly on the case with unlabeled query sets. However, model specialization for deployments isn't limited to the domains we include in our benchmark. We plan to expand this benchmark to capture more scientific domains, including histopathology disease prediction, medical eICU record mortality prediction, satellite imagery for crop classification, and astrophysics galaxy classification.




%
\newpage
\bibliographystyle{ieeenat_fullname}
\bibliography{main}

\begin{thebibliography}{89}
\providecommand{\natexlab}[1]{#1}
\providecommand{\url}[1]{\texttt{#1}}
\expandafter\ifx\csname urlstyle\endcsname\relax
  \providecommand{\doi}[1]{doi: #1}\else
  \providecommand{\doi}{doi: \begingroup \urlstyle{rm}\Url}\fi

\bibitem[Akkaynak and Treibitz(2019)]{akkaynak2019sea}
Derya Akkaynak and Tali Treibitz.
\newblock Sea-thru: A method for removing water from underwater images.
\newblock In \emph{Proceedings of the IEEE/CVF conference on computer vision and pattern recognition}, pages 1682--1691, 2019.

\bibitem[Anonymous(2023)]{anonymous2023when}
Anonymous.
\newblock When less is more: Investigating data pruning for pretraining {LLM}s at scale.
\newblock In \emph{NeurIPS Workshop on Attributing Model Behavior at Scale}, 2023.

\bibitem[Asano et~al.(2020)Asano, Hebisawa, Ishiguro, Takayanagi, Nakamura, Suzuki, Okada, Tanaka, Fukutomi, Ueki, Fukunaga, Konno, Matsuse, Kamei, Taniguchi, Shimoda, and Oguma]{Asano2020NewCD}
Koichiro Asano, Akira Hebisawa, Takashi Ishiguro, Noboru Takayanagi, Yasuhiko Nakamura, Junko Suzuki, Naoki Okada, Jun Tanaka, Yuma Fukutomi, Shigeharu Ueki, Koichi Fukunaga, Satoshi Konno, Hiroto Matsuse, Katsuhiko Kamei, Masami Taniguchi, Terufumi Shimoda, and Tsuyoshi Oguma.
\newblock New clinical diagnostic criteria for allergic bronchopulmonary aspergillosis/mycosis and its validation.
\newblock \emph{The Journal of allergy and clinical immunology}, 2020.

\bibitem[Badgeley et~al.(2018)Badgeley, Zech, Oakden-Rayner, Glicksberg, Liu, Gale, McConnell, Percha, Snyder, and Dudley]{Badgeley2018DeepLP}
Marcus~A. Badgeley, John~R. Zech, Luke Oakden-Rayner, Benjamin~Scott Glicksberg, Manway Liu, William Gale, Michael~V. McConnell, Bethany Percha, Thomas~M. Snyder, and Joel~T. Dudley.
\newblock Deep learning predicts hip fracture using confounding patient and healthcare variables.
\newblock \emph{NPJ Digital Medicine}, 2, 2018.

\bibitem[Bane et~al.(2022)Bane, Uguet, Stribi{\.z}ew, and Zaretskaya]{bane-etal-2022-comparison}
Fred Bane, Celia~Soler Uguet, Wiktor Stribi{\.z}ew, and Anna Zaretskaya.
\newblock A comparison of data filtering methods for neural machine translation.
\newblock In \emph{Proceedings of the 15th Biennial Conference of the Association for Machine Translation in the Americas (Volume 2: Users and Providers Track and Government Track)}, pages 313--325, Orlando, USA, 2022. Association for Machine Translation in the Americas.

\bibitem[Baykal et~al.(2022)Baykal, Liebenwein, Gilitschenski, Feldman, and Rus]{BaykalLGFR22}
Cenk Baykal, Lucas Liebenwein, Igor Gilitschenski, Dan Feldman, and Daniela Rus.
\newblock Sensitivity-informed provable pruning of neural networks.
\newblock \emph{{SIAM} J. Math. Data Sci.}, 4\penalty0 (1):\penalty0 26--45, 2022.

\bibitem[Beery et~al.(2019)Beery, Morris, and Yang]{beery2019efficientpipelinecameratrap}
Sara Beery, Dan Morris, and Siyu Yang.
\newblock Efficient pipeline for camera trap image review, 2019.

\bibitem[Beery et~al.(2021)Beery, Agarwal, Cole, and Birodkar]{beery2021iwildcam}
Sara Beery, Arushi Agarwal, Elijah Cole, and Vighnesh Birodkar.
\newblock The iwildcam 2021 competition dataset, 2021.

\bibitem[Beery et~al.(2022{\natexlab{a}})Beery, Wu, Edwards, Pavetic, Majewski, Mukherjee, Chan, Morgan, Rathod, and Huang]{Beery2022TheAA}
Sara Beery, Guanhang Wu, Trevor Edwards, Filip Pavetic, Bohdan~S. Majewski, Shreyasee Mukherjee, Stanley Chan, John Morgan, Vivek Rathod, and Jonathan Huang.
\newblock The auto arborist dataset: A large-scale benchmark for multiview urban forest monitoring under domain shift.
\newblock \emph{2022 IEEE/CVF Conference on Computer Vision and Pattern Recognition (CVPR)}, pages 21262--21275, 2022{\natexlab{a}}.

\bibitem[Beery et~al.(2022{\natexlab{b}})Beery, Wu, Edwards, Paveti{\'c}, Majewski, Mukherjee, Chan, Morgan, Rathod, and Huang]{beery2022auto}
Sara~Meghan Beery, Guanhang Wu, Trevor Edwards, Filip Paveti{\'c}, Bo Majewski, Shreyasee Mukherjee, Stan Chan, John Morgan, Vivek~Mansing Rathod, and Jonathan Chung-kuan Huang.
\newblock The auto-arborist dataset: A large-scale benchmark for generalizable, multimodal urban forest monitoring.
\newblock 2022{\natexlab{b}}.

\bibitem[Biderman et~al.(2023)Biderman, Schoelkopf, Anthony, Bradley, O'Brien, Hallahan, Khan, Purohit, Prashanth, Raff, Skowron, Sutawika, and van~der Wal]{biderman2023pythia}
Stella Biderman, Hailey Schoelkopf, Quentin Anthony, Herbie Bradley, Kyle O'Brien, Eric Hallahan, Mohammad~Aflah Khan, Shivanshu Purohit, USVSN~Sai Prashanth, Edward Raff, Aviya Skowron, Lintang Sutawika, and Oskar van~der Wal.
\newblock Pythia: A suite for analyzing large language models across training and scaling, 2023.

\bibitem[Black et~al.(2022)Black, Biderman, Hallahan, Anthony, Gao, Golding, He, Leahy, McDonell, Phang, Pieler, Prashanth, Purohit, Reynolds, Tow, Wang, and Weinbach]{black-etal-2022-gpt}
Sidney Black, Stella Biderman, Eric Hallahan, Quentin Anthony, Leo Gao, Laurence Golding, Horace He, Connor Leahy, Kyle McDonell, Jason Phang, Michael Pieler, Usvsn~Sai Prashanth, Shivanshu Purohit, Laria Reynolds, Jonathan Tow, Ben Wang, and Samuel Weinbach.
\newblock {GPT}-{N}eo{X}-20{B}: An open-source autoregressive language model.
\newblock In \emph{Proceedings of BigScience Episode {\#}5 -- Workshop on Challenges {\&} Perspectives in Creating Large Language Models}, pages 95--136, virtual+Dublin, 2022. Association for Computational Linguistics.

\bibitem[Borremans et~al.(2024)Borremans, Durden, Schoening, Curtis, Adams, Albu, Arnaubec, Ayata, Baburaj, Bassin, et~al.]{borremans2024report}
Catherine Borremans, Jennifer Durden, Timm Schoening, Emma Curtis, Luther Adams, Alexandra~Branzan Albu, Aur{\'e}lien Arnaubec, Sakina-Doroth{\'e}e Ayata, Reshma Baburaj, Corinne Bassin, et~al.
\newblock Report on the marine imaging workshop 2022.
\newblock \emph{Research Ideas and Outcomes}, 10:\penalty0 e119782, 2024.

\bibitem[Brandt et~al.(2016)Brandt, Lewis, Fahey, Scott, Darling, and Swanston]{Brandt2016AFF}
Leslie~A. Brandt, Abigail~Derby Lewis, Robert~T. Fahey, Lydia Scott, Lindsay~E. Darling, and Christopher~W. Swanston.
\newblock A framework for adapting urban forests to climate change.
\newblock \emph{Environmental Science \& Policy}, 66:\penalty0 393--402, 2016.

\bibitem[Braverman et~al.(2016)Braverman, Feldman, and Lang]{braverman2016new}
Vladimir Braverman, Dan Feldman, and Harry Lang.
\newblock New frameworks for offline and streaming coreset constructions.
\newblock \emph{arXiv preprint arXiv:1612.00889}, 2016.

\bibitem[Braverman et~al.(2020)Braverman, Drineas, Musco, Musco, Upadhyay, Woodruff, and Zhou]{BravermanDMMUWZ20}
Vladimir Braverman, Petros Drineas, Cameron Musco, Christopher Musco, Jalaj Upadhyay, David~P. Woodruff, and Samson Zhou.
\newblock Near optimal linear algebra in the online and sliding window models.
\newblock In \emph{61st {IEEE} Annual Symposium on Foundations of Computer Science, {FOCS}}, pages 517--528, 2020.

\bibitem[Brown et~al.(2020)Brown, Mann, Ryder, Subbiah, Kaplan, Dhariwal, Neelakantan, Shyam, Sastry, Askell, Agarwal, Herbert-Voss, Krueger, Henighan, Child, Ramesh, Ziegler, Wu, Winter, Hesse, Chen, Sigler, teusz Litwin, Gray, Chess, Clark, Berner, McCandlish, Radford, Sutskever, and Amodei]{Brown2020LanguageMA}
Tom~B. Brown, Benjamin Mann, Nick Ryder, Melanie Subbiah, Jared Kaplan, Prafulla Dhariwal, Arvind Neelakantan, Pranav Shyam, Girish Sastry, Amanda Askell, Sandhini Agarwal, Ariel Herbert-Voss, Gretchen Krueger, Tom Henighan, Rewon Child, Aditya Ramesh, Daniel~M. Ziegler, Jeff Wu, Clemens Winter, Christopher Hesse, Mark Chen, Eric Sigler, Ma teusz Litwin, Scott Gray, Benjamin Chess, Jack Clark, Christopher Berner, Sam McCandlish, Alec Radford, Ilya Sutskever, and Dario Amodei.
\newblock Language models are few-shot learners.
\newblock \emph{ArXiv}, abs/2005.14165, 2020.

\bibitem[Caesar et~al.(2020)Caesar, Bankiti, Lang, Vora, Liong, Xu, Krishnan, Pan, Baldan, and Beijbom]{caesar2020nuscenes}
Holger Caesar, Varun Bankiti, Alex~H Lang, Sourabh Vora, Venice~Erin Liong, Qiang Xu, Anush Krishnan, Yu Pan, Giancarlo Baldan, and Oscar Beijbom.
\newblock nuscenes: A multimodal dataset for autonomous driving.
\newblock In \emph{Proceedings of the IEEE/CVF conference on computer vision and pattern recognition}, pages 11621--11631, 2020.

\bibitem[Chen(2009)]{Chen09}
Ke Chen.
\newblock On coresets for k-median and k-means clustering in metric and euclidean spaces and their applications.
\newblock \emph{{SIAM} J. Comput.}, 39\penalty0 (3):\penalty0 923--947, 2009.

\bibitem[Chhaya et~al.(2020)Chhaya, Dasgupta, and Shit]{Chhaya0S20}
Rachit Chhaya, Anirban Dasgupta, and Supratim Shit.
\newblock On coresets for regularized regression.
\newblock In \emph{Proceedings of the 37th International Conference on Machine Learning, {ICML}}, 2020.

\bibitem[Clarkson(2010)]{Clarkson10}
Kenneth~L. Clarkson.
\newblock Coresets, sparse greedy approximation, and the frank-wolfe algorithm.
\newblock \emph{{ACM} Trans. Algorithms}, 6\penalty0 (4):\penalty0 63:1--63:30, 2010.

\bibitem[Cohen et~al.(2017)Cohen, Musco, and Musco]{CohenMM17}
Michael~B. Cohen, Cameron Musco, and Christopher Musco.
\newblock Input sparsity time low-rank approximation via ridge leverage score sampling.
\newblock In \emph{Proceedings of the Twenty-Eighth Annual {ACM-SIAM} Symposium on Discrete Algorithms, {SODA}}, pages 1758--1777, 2017.

\bibitem[Cohen-Addad et~al.(2022)Cohen-Addad, Green~Larsen, Saulpic, Schwiegelshohn, and Sheikh-Omar]{cohen2022improved}
Vincent Cohen-Addad, Kasper Green~Larsen, David Saulpic, Chris Schwiegelshohn, and Omar~Ali Sheikh-Omar.
\newblock Improved coresets for euclidean $ k $-means.
\newblock \emph{Advances in Neural Information Processing Systems}, 35:\penalty0 2679--2694, 2022.

\bibitem[Coleman et~al.(2019)Coleman, Yeh, Mussmann, Mirzasoleiman, Bailis, Liang, Leskovec, and Zaharia]{coleman2019selection}
Cody Coleman, Christopher Yeh, Stephen Mussmann, Baharan Mirzasoleiman, Peter Bailis, Percy Liang, Jure Leskovec, and Matei Zaharia.
\newblock Selection via proxy: Efficient data selection for deep learning.
\newblock In \emph{International Conference on Learning Representations}, 2019.

\bibitem[Compton et~al.(2023)Compton, Zhang, Puli, and Ranganath]{Compton2023WhenMI}
Rhys Compton, Lily~H. Zhang, Aahlad~Manas Puli, and Rajesh Ranganath.
\newblock When more is less: Incorporating additional datasets can hurt performance by introducing spurious correlations.
\newblock \emph{ArXiv}, abs/2308.04431, 2023.

\bibitem[Dasgupta et~al.(2008)Dasgupta, Drineas, Harb, Kumar, and Mahoney]{DasguptaDHKM08}
Anirban Dasgupta, Petros Drineas, Boulos Harb, Ravi Kumar, and Michael~W. Mahoney.
\newblock Sampling algorithms and coresets for $l_p$ regression.
\newblock In \emph{Proceedings of the Nineteenth Annual {ACM-SIAM} Symposium on Discrete Algorithms, {SODA}}, pages 932--941, 2008.

\bibitem[Dawkins et~al.(2017)Dawkins, Sherrill, Fieldhouse, Hoogs, Richards, Zhang, Prasad, Williams, Lauffenburger, and Wang]{viame}
Matthew Dawkins, Linus Sherrill, Keith Fieldhouse, Anthony Hoogs, Benjamin Richards, David Zhang, Lakshman Prasad, Kresimir Williams, Nathan Lauffenburger, and Gaoang Wang.
\newblock An open-source platform for underwater image and video analytics.
\newblock In \emph{2017 IEEE Winter Conference on Applications of Computer Vision (WACV)}, pages 898--906, 2017.

\bibitem[Deng et~al.(2009{\natexlab{a}})Deng, Dong, Socher, Li, Li, and Fei-Fei]{imagenet}
Jia Deng, Wei Dong, Richard Socher, Li-Jia Li, Kai Li, and Li Fei-Fei.
\newblock Imagenet: A large-scale hierarchical image database.
\newblock In \emph{2009 IEEE Conference on Computer Vision and Pattern Recognition}, pages 248--255, 2009{\natexlab{a}}.

\bibitem[Deng et~al.(2009{\natexlab{b}})Deng, Dong, Socher, Li, Li, and Fei-Fei]{imgnet}
Jia Deng, Wei Dong, Richard Socher, Li-Jia Li, Kai Li, and Li Fei-Fei.
\newblock Imagenet: A large-scale hierarchical image database.
\newblock In \emph{2009 IEEE Conference on Computer Vision and Pattern Recognition}, pages 248--255, 2009{\natexlab{b}}.

\bibitem[Di~Lorenzo et~al.(2022)Di~Lorenzo, L{\o}nborg, Andersen, Escobar-Briones, Devlin, Borja, M{\"u}ller, Robinson, Ford, Zivian, et~al.]{di2022sustainable}
Emanuele Di~Lorenzo, Christian L{\o}nborg, Jesper~H Andersen, Elva~G Escobar-Briones, Michelle~Jillian Devlin, Angel Borja, Marius~Nils M{\"u}ller, Carol Robinson, Alex Ford, Anna~Milena Zivian, et~al.
\newblock \emph{Sustainable Development Goal 14-Life Below Water: Towards a Sustainable Ocean}.
\newblock Frontiers Media SA, 2022.

\bibitem[Dosovitskiy et~al.(2020)Dosovitskiy, Beyer, Kolesnikov, Weissenborn, Zhai, Unterthiner, Dehghani, Minderer, Heigold, Gelly, Uszkoreit, and Houlsby]{Dosovitskiy2020AnII}
Alexey Dosovitskiy, Lucas Beyer, Alexander Kolesnikov, Dirk Weissenborn, Xiaohua Zhai, Thomas Unterthiner, Mostafa Dehghani, Matthias Minderer, Georg Heigold, Sylvain Gelly, Jakob Uszkoreit, and Neil Houlsby.
\newblock An image is worth 16x16 words: Transformers for image recognition at scale.
\newblock \emph{ArXiv}, abs/2010.11929, 2020.

\bibitem[Ein-Dor et~al.(2020)Ein-Dor, Halfon, Gera, Shnarch, Dankin, Choshen, Danilevsky, Aharonov, Katz, and Slonim]{ein-dor-etal-2020-active}
Liat Ein-Dor, Alon Halfon, Ariel Gera, Eyal Shnarch, Lena Dankin, Leshem Choshen, Marina Danilevsky, Ranit Aharonov, Yoav Katz, and Noam Slonim.
\newblock {A}ctive {L}earning for {BERT}: {A}n {E}mpirical {S}tudy.
\newblock In \emph{Proceedings of the 2020 Conference on Empirical Methods in Natural Language Processing (EMNLP)}, pages 7949--7962, Online, 2020. Association for Computational Linguistics.

\bibitem[Feuer et~al.(2024)Feuer, Xu, Cohen, Yubeaton, Mittal, and Hegde]{feuer2024select}
Benjamin Feuer, Jiawei Xu, Niv Cohen, Patrick Yubeaton, Govind Mittal, and Chinmay Hegde.
\newblock Select: A large-scale benchmark of data curation strategies for image classification.
\newblock \emph{arXiv preprint arXiv:2410.05057}, 2024.

\bibitem[Gadre et~al.(2024)Gadre, Ilharco, Fang, Hayase, Smyrnis, Nguyen, Marten, Wortsman, Ghosh, Zhang, et~al.]{gadre2024datacomp}
Samir~Yitzhak Gadre, Gabriel Ilharco, Alex Fang, Jonathan Hayase, Georgios Smyrnis, Thao Nguyen, Ryan Marten, Mitchell Wortsman, Dhruba Ghosh, Jieyu Zhang, et~al.
\newblock Datacomp: In search of the next generation of multimodal datasets.
\newblock \emph{Advances in Neural Information Processing Systems}, 36, 2024.

\bibitem[Geirhos et~al.(2020)Geirhos, Jacobsen, Michaelis, Zemel, Brendel, Bethge, and Wichmann]{Geirhos2020ShortcutLI}
Robert Geirhos, J{\"o}rn-Henrik Jacobsen, Claudio Michaelis, Richard~S. Zemel, Wieland Brendel, Matthias Bethge, and Felix Wichmann.
\newblock Shortcut learning in deep neural networks.
\newblock \emph{Nature Machine Intelligence}, 2:\penalty0 665 -- 673, 2020.

\bibitem[Har{-}Peled and Mazumdar(2004)]{Har-PeledM04}
Sariel Har{-}Peled and Soham Mazumdar.
\newblock On coresets for k-means and k-median clustering.
\newblock In \emph{Proceedings of the 36th Annual {ACM} Symposium on Theory of Computing}, pages 291--300, 2004.

\bibitem[He et~al.(2015)He, Zhang, Ren, and Sun]{He2015DeepRL}
Kaiming He, X. Zhang, Shaoqing Ren, and Jian Sun.
\newblock Deep residual learning for image recognition.
\newblock \emph{2016 IEEE Conference on Computer Vision and Pattern Recognition (CVPR)}, pages 770--778, 2015.

\bibitem[Huang and Vishnoi(2020)]{HuangV20}
Lingxiao Huang and Nisheeth~K. Vishnoi.
\newblock Coresets for clustering in euclidean spaces: importance sampling is nearly optimal.
\newblock In \emph{Proccedings of the 52nd Annual {ACM} {SIGACT} Symposium on Theory of Computing, {STOC}}, pages 1416--1429, 2020.

\bibitem[Jerlov(1976)]{jerlov1976marine}
Nils~Gunnar Jerlov.
\newblock \emph{Marine optics}.
\newblock Elsevier, 1976.

\bibitem[Jocher et~al.(2023)Jocher, Qiu, and Chaurasia]{Jocher_Ultralytics_YOLO_2023}
Glenn Jocher, Jing Qiu, and Ayush Chaurasia.
\newblock {Ultralytics YOLO}, 2023.

\bibitem[Johnson et~al.(2019)Johnson, Douze, and J{\'e}gou]{faiss_gpu}
Jeff Johnson, Matthijs Douze, and Herv{\'e} J{\'e}gou.
\newblock Billion-scale similarity search with {GPUs}.
\newblock \emph{IEEE Transactions on Big Data}, 7\penalty0 (3):\penalty0 535--547, 2019.

\bibitem[Jubran et~al.(2020)Jubran, Tukan, Maalouf, and Feldman]{jubran2020sets}
Ibrahim Jubran, Murad Tukan, Alaa Maalouf, and Dan Feldman.
\newblock Sets clustering.
\newblock In \emph{International Conference on Machine Learning}, pages 4994--5005. PMLR, 2020.

\bibitem[Killamsetty et~al.(2021{\natexlab{a}})Killamsetty, Sivasubramanian, Ramakrishnan, De, and Iyer]{killamsetty2021grad}
KrishnaTeja Killamsetty, Durga Sivasubramanian, Ganesh Ramakrishnan, Abir De, and Rishabh~K. Iyer.
\newblock {GRAD-MATCH:} gradient matching based data subset selection for efficient deep model training.
\newblock In \emph{Proceedings of the 38th International Conference on Machine Learning, {ICML}}, pages 5464--5474, 2021{\natexlab{a}}.

\bibitem[Killamsetty et~al.(2021{\natexlab{b}})Killamsetty, Sivasubramanian, Ramakrishnan, and Iyer]{killamsetty2021glister}
KrishnaTeja Killamsetty, Durga Sivasubramanian, Ganesh Ramakrishnan, and Rishabh~K. Iyer.
\newblock {GLISTER:} generalization based data subset selection for efficient and robust learning.
\newblock In \emph{Thirty-Fifth {AAAI} Conference on Artificial Intelligence, {AAAI}}, 2021{\natexlab{b}}.

\bibitem[Killamsetty et~al.(2021{\natexlab{c}})Killamsetty, Zhao, Chen, and Iyer]{killamsetty2021retrieve}
Krishnateja Killamsetty, Xujiang Zhao, Feng Chen, and Rishabh Iyer.
\newblock Retrieve: Coreset selection for efficient and robust semi-supervised learning.
\newblock In \emph{Advances in Neural Information Processing Systems (NeurIPS)}, 2021{\natexlab{c}}.

\bibitem[Koh et~al.(2020)Koh, Sagawa, Marklund, Xie, Zhang, Balsubramani, Hu, Yasunaga, Phillips, Gao, Lee, David, Stavness, Guo, Earnshaw, Haque, Beery, Leskovec, Kundaje, Pierson, Levine, Finn, and Liang]{Koh2020WILDSAB}
Pang~Wei Koh, Shiori Sagawa, Henrik Marklund, Sang~Michael Xie, Marvin Zhang, Akshay Balsubramani, Weihua Hu, Michihiro Yasunaga, Richard~Lanas Phillips, Irena Gao, Tony Lee, Etiene David, Ian Stavness, Wei Guo, Berton~A. Earnshaw, Imran~S. Haque, Sara~Meghan Beery, Jure Leskovec, Anshul Kundaje, Emma Pierson, Sergey Levine, Chelsea Finn, and Percy Liang.
\newblock Wilds: A benchmark of in-the-wild distribution shifts.
\newblock In \emph{International Conference on Machine Learning}, 2020.

\bibitem[Krizhevsky et~al.(2009)Krizhevsky, Hinton, et~al.]{krizhevsky2009learning}
Alex Krizhevsky, Geoffrey Hinton, et~al.
\newblock Learning multiple layers of features from tiny images, 2009.

\bibitem[Kuznetsova et~al.(2018)Kuznetsova, Rom, Alldrin, Uijlings, Krasin, Pont-Tuset, Kamali, Popov, Malloci, Kolesnikov, Duerig, and Ferrari]{Kuznetsova2018TheOI}
Alina Kuznetsova, Hassan Rom, Neil~Gordon Alldrin, Jasper R.~R. Uijlings, Ivan Krasin, Jordi Pont-Tuset, Shahab Kamali, Stefan Popov, Matteo Malloci, Alexander Kolesnikov, Tom Duerig, and Vittorio Ferrari.
\newblock The open images dataset v4.
\newblock \emph{International Journal of Computer Vision}, 128:\penalty0 1956 -- 1981, 2018.

\bibitem[Laurençon et~al.(2023)Laurençon, Saulnier, Wang, Akiki, del Moral, Scao, Werra, Mou, Ponferrada, Nguyen, Frohberg, Šaško, Lhoest, McMillan-Major, Dupont, Biderman, Rogers, allal, Toni, Pistilli, Nguyen, Nikpoor, Masoud, Colombo, de~la Rosa, Villegas, Thrush, Longpre, Nagel, Weber, Muñoz, Zhu, Strien, Alyafeai, Almubarak, Vu, Gonzalez-Dios, Soroa, Lo, Dey, Suarez, Gokaslan, Bose, Adelani, Phan, Tran, Yu, Pai, Chim, Lepercq, Ilic, Mitchell, Luccioni, and Jernite]{laurençon2023bigscience}
Hugo Laurençon, Lucile Saulnier, Thomas Wang, Christopher Akiki, Albert~Villanova del Moral, Teven~Le Scao, Leandro~Von Werra, Chenghao Mou, Eduardo~González Ponferrada, Huu Nguyen, Jörg Frohberg, Mario Šaško, Quentin Lhoest, Angelina McMillan-Major, Gerard Dupont, Stella Biderman, Anna Rogers, Loubna~Ben allal, Francesco~De Toni, Giada Pistilli, Olivier Nguyen, Somaieh Nikpoor, Maraim Masoud, Pierre Colombo, Javier de~la Rosa, Paulo Villegas, Tristan Thrush, Shayne Longpre, Sebastian Nagel, Leon Weber, Manuel Muñoz, Jian Zhu, Daniel~Van Strien, Zaid Alyafeai, Khalid Almubarak, Minh~Chien Vu, Itziar Gonzalez-Dios, Aitor Soroa, Kyle Lo, Manan Dey, Pedro~Ortiz Suarez, Aaron Gokaslan, Shamik Bose, David Adelani, Long Phan, Hieu Tran, Ian Yu, Suhas Pai, Jenny Chim, Violette Lepercq, Suzana Ilic, Margaret Mitchell, Sasha~Alexandra Luccioni, and Yacine Jernite.
\newblock The bigscience roots corpus: A 1.6tb composite multilingual dataset, 2023.

\bibitem[Liebenwein et~al.(2019)Liebenwein, Baykal, Lang, Feldman, and Rus]{liebenwein2019provable}
Lucas Liebenwein, Cenk Baykal, Harry Lang, Dan Feldman, and Daniela Rus.
\newblock Provable filter pruning for efficient neural networks.
\newblock In \emph{International Conference on Learning Representations}, 2019.

\bibitem[Maalouf et~al.(2019)Maalouf, Jubran, and Feldman]{maalouf2019fast}
Alaa Maalouf, Ibrahim Jubran, and Dan Feldman.
\newblock Fast and accurate least-mean-squares solvers.
\newblock In \emph{Proceedings of the 33rd International Conference on Neural Information Processing Systems}, pages 8307--8318, 2019.

\bibitem[Maalouf et~al.(2021)Maalouf, Jubran, Tukan, and Feldman]{maalouf2021coresets}
Alaa Maalouf, Ibrahim Jubran, Murad Tukan, and Dan Feldman.
\newblock Coresets for the average case error for finite query sets.
\newblock \emph{Sensors}, 21\penalty0 (19):\penalty0 6689, 2021.

\bibitem[Maalouf et~al.(2022)Maalouf, Eini, Mussay, Feldman, and Osadchy]{maalouf2022unified}
Alaa Maalouf, Gilad Eini, Ben Mussay, Dan Feldman, and Margarita Osadchy.
\newblock A unified approach to coreset learning.
\newblock \emph{IEEE Transactions on Neural Networks and Learning Systems}, 2022.

\bibitem[Mazumder et~al.(2023)Mazumder, Banbury, Yao, Karlaš, Rojas, Diamos, Diamos, He, Parrish, Kirk, Quaye, Rastogi, Kiela, Jurado, Kanter, Mosquera, Ciro, Aroyo, Acun, Chen, Raje, Bartolo, Eyuboglu, Ghorbani, Goodman, Inel, Kane, Kirkpatrick, Kuo, Mueller, Thrush, Vanschoren, Warren, Williams, Yeung, Ardalani, Paritosh, Bat-Leah, Zhang, Zou, Wu, Coleman, Ng, Mattson, and Reddi]{mazumder2023dataperfbenchmarksdatacentricai}
Mark Mazumder, Colby Banbury, Xiaozhe Yao, Bojan Karlaš, William~Gaviria Rojas, Sudnya Diamos, Greg Diamos, Lynn He, Alicia Parrish, Hannah~Rose Kirk, Jessica Quaye, Charvi Rastogi, Douwe Kiela, David Jurado, David Kanter, Rafael Mosquera, Juan Ciro, Lora Aroyo, Bilge Acun, Lingjiao Chen, Mehul~Smriti Raje, Max Bartolo, Sabri Eyuboglu, Amirata Ghorbani, Emmett Goodman, Oana Inel, Tariq Kane, Christine~R. Kirkpatrick, Tzu-Sheng Kuo, Jonas Mueller, Tristan Thrush, Joaquin Vanschoren, Margaret Warren, Adina Williams, Serena Yeung, Newsha Ardalani, Praveen Paritosh, Lilith Bat-Leah, Ce Zhang, James Zou, Carole-Jean Wu, Cody Coleman, Andrew Ng, Peter Mattson, and Vijay~Janapa Reddi.
\newblock Dataperf: Benchmarks for data-centric ai development, 2023.

\bibitem[Meyer et~al.(2022)Meyer, Musco, Musco, Woodruff, and Zhou]{MeyerMMWZ22}
Raphael~A. Meyer, Cameron Musco, Christopher Musco, David~P. Woodruff, and Samson Zhou.
\newblock Fast regression for structured inputs.
\newblock In \emph{The Tenth International Conference on Learning Representations, {ICLR}}, 2022.

\bibitem[Mindermann et~al.(2022)Mindermann, Brauner, Razzak, Sharma, Kirsch, Xu, Höltgen, Gomez, Morisot, Farquhar, and Gal]{mindermann2022prioritized}
Sören Mindermann, Jan Brauner, Muhammed Razzak, Mrinank Sharma, Andreas Kirsch, Winnie Xu, Benedikt Höltgen, Aidan~N. Gomez, Adrien Morisot, Sebastian Farquhar, and Yarin Gal.
\newblock Prioritized training on points that are learnable, worth learning, and not yet learnt, 2022.

\bibitem[Mirzasoleiman et~al.(2020{\natexlab{a}})Mirzasoleiman, Bilmes, and Leskovec]{mirzasoleiman2020coresets}
Baharan Mirzasoleiman, Jeff~A. Bilmes, and Jure Leskovec.
\newblock Coresets for data-efficient training of machine learning models.
\newblock In \emph{Proceedings of the 37th International Conference on Machine Learning, {ICML}}, pages 6950--6960, 2020{\natexlab{a}}.

\bibitem[Mirzasoleiman et~al.(2020{\natexlab{b}})Mirzasoleiman, Cao, and Leskovec]{MirzasoleimanCL20}
Baharan Mirzasoleiman, Kaidi Cao, and Jure Leskovec.
\newblock Coresets for robust training of deep neural networks against noisy labels.
\newblock In \emph{Advances in Neural Information Processing Systems 33: Annual Conference on Neural Information Processing Systems, NeurIPS}, 2020{\natexlab{b}}.

\bibitem[Muennighoff et~al.(2023)Muennighoff, Rush, Barak, Scao, Piktus, Tazi, Pyysalo, Wolf, and Raffel]{muennighoff2023scaling}
Niklas Muennighoff, Alexander~M. Rush, Boaz Barak, Teven~Le Scao, Aleksandra Piktus, Nouamane Tazi, Sampo Pyysalo, Thomas Wolf, and Colin Raffel.
\newblock Scaling data-constrained language models, 2023.

\bibitem[Norouzzadeh et~al.(2019)Norouzzadeh, Morris, Beery, Joshi, Jojic, and Clune]{norouzzadeh2019deepactivelearningspecies}
Mohammad~Sadegh Norouzzadeh, Dan Morris, Sara Beery, Neel Joshi, Nebojsa Jojic, and Jeff Clune.
\newblock A deep active learning system for species identification and counting in camera trap images, 2019.

\bibitem[Paul et~al.(2021)Paul, Ganguli, and Dziugaite]{paul2021diet}
Mansheej Paul, Surya Ganguli, and Gintare~Karolina Dziugaite.
\newblock Deep learning on a data diet: Finding important examples early in training.
\newblock In \emph{Association for the Advancement of Artificial Intelligence (AAAI)}, 2021.

\bibitem[Penedo et~al.(2023)Penedo, Malartic, Hesslow, Cojocaru, Cappelli, Alobeidli, Pannier, Almazrouei, and Launay]{penedo2023refinedweb}
Guilherme Penedo, Quentin Malartic, Daniel Hesslow, Ruxandra Cojocaru, Alessandro Cappelli, Hamza Alobeidli, Baptiste Pannier, Ebtesam Almazrouei, and Julien Launay.
\newblock The refinedweb dataset for falcon llm: Outperforming curated corpora with web data, and web data only, 2023.

\bibitem[Rae et~al.(2022)Rae, Borgeaud, Cai, Millican, Hoffmann, Song, Aslanides, Henderson, Ring, Young, Rutherford, Hennigan, Menick, Cassirer, Powell, van~den Driessche, Hendricks, Rauh, Huang, Glaese, Welbl, Dathathri, Huang, Uesato, Mellor, Higgins, Creswell, McAleese, Wu, Elsen, Jayakumar, Buchatskaya, Budden, Sutherland, Simonyan, Paganini, Sifre, Martens, Li, Kuncoro, Nematzadeh, Gribovskaya, Donato, Lazaridou, Mensch, Lespiau, Tsimpoukelli, Grigorev, Fritz, Sottiaux, Pajarskas, Pohlen, Gong, Toyama, de~Masson~d'Autume, Li, Terzi, Mikulik, Babuschkin, Clark, de~Las~Casas, Guy, Jones, Bradbury, Johnson, Hechtman, Weidinger, Gabriel, Isaac, Lockhart, Osindero, Rimell, Dyer, Vinyals, Ayoub, Stanway, Bennett, Hassabis, Kavukcuoglu, and Irving]{rae2022scaling}
Jack~W. Rae, Sebastian Borgeaud, Trevor Cai, Katie Millican, Jordan Hoffmann, Francis Song, John Aslanides, Sarah Henderson, Roman Ring, Susannah Young, Eliza Rutherford, Tom Hennigan, Jacob Menick, Albin Cassirer, Richard Powell, George van~den Driessche, Lisa~Anne Hendricks, Maribeth Rauh, Po-Sen Huang, Amelia Glaese, Johannes Welbl, Sumanth Dathathri, Saffron Huang, Jonathan Uesato, John Mellor, Irina Higgins, Antonia Creswell, Nat McAleese, Amy Wu, Erich Elsen, Siddhant Jayakumar, Elena Buchatskaya, David Budden, Esme Sutherland, Karen Simonyan, Michela Paganini, Laurent Sifre, Lena Martens, Xiang~Lorraine Li, Adhiguna Kuncoro, Aida Nematzadeh, Elena Gribovskaya, Domenic Donato, Angeliki Lazaridou, Arthur Mensch, Jean-Baptiste Lespiau, Maria Tsimpoukelli, Nikolai Grigorev, Doug Fritz, Thibault Sottiaux, Mantas Pajarskas, Toby Pohlen, Zhitao Gong, Daniel Toyama, Cyprien de Masson~d'Autume, Yujia Li, Tayfun Terzi, Vladimir Mikulik, Igor Babuschkin, Aidan Clark, Diego de Las~Casas, Aurelia Guy, Chris Jones,
  James Bradbury, Matthew Johnson, Blake Hechtman, Laura Weidinger, Iason Gabriel, William Isaac, Ed Lockhart, Simon Osindero, Laura Rimell, Chris Dyer, Oriol Vinyals, Kareem Ayoub, Jeff Stanway, Lorrayne Bennett, Demis Hassabis, Koray Kavukcuoglu, and Geoffrey Irving.
\newblock Scaling language models: Methods, analysis and insights from training gopher, 2022.

\bibitem[Raffel et~al.(2020)Raffel, Shazeer, Roberts, Lee, Narang, Matena, Zhou, Li, and Liu]{raffel2020exploring}
Colin Raffel, Noam Shazeer, Adam Roberts, Katherine Lee, Sharan Narang, Michael Matena, Yanqi Zhou, Wei Li, and Peter~J. Liu.
\newblock Exploring the limits of transfer learning with a unified text-to-text transformer, 2020.

\bibitem[Ramaswamy et~al.(2023)Ramaswamy, Lin, Zhao, Adcock, van~der Maaten, Ghadiyaram, and Russakovsky]{ramaswamy2023geode}
Vikram~V. Ramaswamy, Sing~Yu Lin, Dora Zhao, Aaron~B. Adcock, Laurens van~der Maaten, Deepti Ghadiyaram, and Olga Russakovsky.
\newblock Geode: a geographically diverse evaluation dataset for object recognition, 2023.

\bibitem[Russakovsky et~al.(2014)Russakovsky, Deng, Su, Krause, Satheesh, Ma, Huang, Karpathy, Khosla, Bernstein, Berg, and Fei-Fei]{Russakovsky2014ImageNetLS}
Olga Russakovsky, Jia Deng, Hao Su, Jonathan Krause, Sanjeev Satheesh, Sean Ma, Zhiheng Huang, Andrej Karpathy, Aditya Khosla, Michael~S. Bernstein, Alexander~C. Berg, and Li Fei-Fei.
\newblock Imagenet large scale visual recognition challenge.
\newblock \emph{International Journal of Computer Vision}, 115:\penalty0 211 -- 252, 2014.

\bibitem[Sener and Savarese(2018)]{sener2018active}
Ozan Sener and Silvio Savarese.
\newblock Active learning for convolutional neural networks: A core-set approach.
\newblock In \emph{International Conference on Learning Representations (ICLR)}, 2018.

\bibitem[Shankar et~al.(2017)Shankar, Halpern, Breck, Atwood, Wilson, and Sculley]{Shankar2017NoCW}
Shreya Shankar, Yoni Halpern, Eric Breck, James Atwood, Jimbo Wilson, and D. Sculley.
\newblock No classification without representation: Assessing geodiversity issues in open data sets for the developing world.
\newblock \emph{arXiv: Machine Learning}, 2017.

\bibitem[Shen et~al.(2024)Shen, Raji, and Chen]{shen2024dataadditiondilemma}
Judy~Hanwen Shen, Inioluwa~Deborah Raji, and Irene~Y. Chen.
\newblock The data addition dilemma, 2024.

\bibitem[Siddiqui et~al.(2022)Siddiqui, Rajkumar, Maharaj, Krueger, and Hooker]{siddiqui2022metadata}
Shoaib~Ahmed Siddiqui, Nitarshan Rajkumar, Tegan Maharaj, David Krueger, and Sara Hooker.
\newblock Metadata archaeology: Unearthing data subsets by leveraging training dynamics, 2022.

\bibitem[Sorscher et~al.(2022)Sorscher, Geirhos, Shekhar, Ganguli, and Morcos]{sorscher2022beyond}
Ben Sorscher, Robert Geirhos, Shashank Shekhar, Surya Ganguli, and Ari~S. Morcos.
\newblock Beyond neural scaling laws: beating power law scaling via data pruning.
\newblock \emph{arXiv}, 2022.

\bibitem[Sorscher et~al.(2023)Sorscher, Geirhos, Shekhar, Ganguli, and Morcos]{sorscher2023neural}
Ben Sorscher, Robert Geirhos, Shashank Shekhar, Surya Ganguli, and Ari~S. Morcos.
\newblock Beyond neural scaling laws: beating power law scaling via data pruning, 2023.

\bibitem[Staub(2020)]{Staub_2020}
Francis Staub.
\newblock Living planet report 2020: Bending the curve of biodiversity loss, 2020.

\bibitem[Tamkin et~al.(2022)Tamkin, Nguyen, Deshpande, Mu, and Goodman]{tamkin2022active}
Alex Tamkin, Dat Nguyen, Salil Deshpande, Jesse Mu, and Noah Goodman.
\newblock Active learning helps pretrained models learn the intended task, 2022.

\bibitem[Taori et~al.(2020)Taori, Dave, Shankar, Carlini, Recht, and Schmidt]{Taori2020MeasuringRT}
Rohan Taori, Achal Dave, Vaishaal Shankar, Nicholas Carlini, Benjamin Recht, and Ludwig Schmidt.
\newblock Measuring robustness to natural distribution shifts in image classification.
\newblock \emph{ArXiv}, abs/2007.00644, 2020.

\bibitem[Tolochinsky et~al.(2022)Tolochinsky, Jubran, and Feldman]{TolochinskyJF22}
Elad Tolochinsky, Ibrahim Jubran, and Dan Feldman.
\newblock Generic coreset for scalable learning of monotonic kernels: Logistic regression, sigmoid and more.
\newblock In \emph{International Conference on Machine Learning, {ICML}}, 2022.

\bibitem[Tukan et~al.(2021)Tukan, Baykal, Feldman, and Rus]{TukanBFR21}
Murad Tukan, Cenk Baykal, Dan Feldman, and Daniela Rus.
\newblock On coresets for support vector machines.
\newblock \emph{Theor. Comput. Sci.}, 890:\penalty0 171--191, 2021.

\bibitem[Tukan et~al.(2022)Tukan, Mualem, and Maalouf]{Tukan2022provable}
Murad Tukan, Loay Mualem, and Alaa Maalouf.
\newblock Pruning neural networks via coresets and convex geometry: Towards no assumptions.
\newblock In \emph{Proceedings of the 36th International Conference on Neural Information Processing Systems}, 2022.

\bibitem[Tukan et~al.(2023)Tukan, Zhou, Maalouf, Rus, Braverman, and Feldman]{tukan2023provable}
Murad Tukan, Samson Zhou, Alaa Maalouf, Daniela Rus, Vladimir Braverman, and Dan Feldman.
\newblock Provable data subset selection for efficient neural networks training.
\newblock In \emph{International Conference on Machine Learning}, pages 34533--34555. PMLR, 2023.

\bibitem[{United Nations}(2023)]{UN_BNJJ}
{United Nations}.
\newblock Agreement under the {United Nations Convention on the Law of the Sea} on the {Conservation and Sustainable Use of Marine Biological Diversity of Areas Beyond National Jurisdiction}.
\newblock United Nations Treaty Collection, 2023.
\newblock Opened for signature on 20 September 2023.

\bibitem[Wang et~al.(2021)Wang, Lan, Liu, Ouyang, and Qin]{Wang2021GeneralizingTU}
Jindong Wang, Cuiling Lan, Chang Liu, Yidong Ouyang, and Tao Qin.
\newblock Generalizing to unseen domains: A survey on domain generalization.
\newblock \emph{IEEE Transactions on Knowledge and Data Engineering}, 35:\penalty0 8052--8072, 2021.

\bibitem[Wang et~al.(2024)Wang, Maalouf, Xiao, Ban, Amini, Rosman, Karaman, and Rus]{wang2024drive}
Tsun-Hsuan Wang, Alaa Maalouf, Wei Xiao, Yutong Ban, Alexander Amini, Guy Rosman, Sertac Karaman, and Daniela Rus.
\newblock Drive anywhere: Generalizable end-to-end autonomous driving with multi-modal foundation models.
\newblock In \emph{2024 IEEE International Conference on Robotics and Automation (ICRA)}, pages 6687--6694. IEEE, 2024.

\bibitem[Wang et~al.(2020)Wang, Pham, Michel, Anastasopoulos, Carbonell, and Neubig]{wang2020optimizing}
Xinyi Wang, Hieu Pham, Paul Michel, Antonios Anastasopoulos, Jaime Carbonell, and Graham Neubig.
\newblock Optimizing data usage via differentiable rewards.
\newblock In \emph{International Conference on Machine Learning (ICML)}, 2020.

\bibitem[Wang et~al.(2022)Wang, Lian, and Yu]{wang2022unsupervised}
Xudong Wang, Long Lian, and Stella~X Yu.
\newblock Unsupervised selective labeling for more effective semi-supervised learning.
\newblock In \emph{European Conference on Computer Vision}, pages 427--445. Springer, 2022.

\bibitem[Wearn and Glover-Kapfer(2017)]{camera_traps_best_practice}
Oliver Wearn and Paul Glover-Kapfer.
\newblock Camera-trapping for conservation: a guide to best-practices, 2017.

\bibitem[Wenzek et~al.(2020)Wenzek, Lachaux, Conneau, Chaudhary, Guzm{\'a}n, Joulin, and Grave]{wenzek-etal-2020-ccnet}
Guillaume Wenzek, Marie-Anne Lachaux, Alexis Conneau, Vishrav Chaudhary, Francisco Guzm{\'a}n, Armand Joulin, and Edouard Grave.
\newblock {CCN}et: Extracting high quality monolingual datasets from web crawl data.
\newblock In \emph{Proceedings of the Twelfth Language Resources and Evaluation Conference}, pages 4003--4012, Marseille, France, 2020. European Language Resources Association.

\bibitem[Yuan et~al.(2020)Yuan, Lin, and Boyd-Graber]{yuan-etal-2020-cold}
Michelle Yuan, Hsuan-Tien Lin, and Jordan Boyd-Graber.
\newblock Cold-start active learning through self-supervised language modeling.
\newblock In \emph{Proceedings of the 2020 Conference on Empirical Methods in Natural Language Processing (EMNLP)}, pages 7935--7948, Online, 2020. Association for Computational Linguistics.

\bibitem[Zhang et~al.(2022)Zhang, Roller, Goyal, Artetxe, Chen, Chen, Dewan, Diab, Li, Lin, Mihaylov, Ott, Shleifer, Shuster, Simig, Koura, Sridhar, Wang, and Zettlemoyer]{zhang2022opt}
Susan Zhang, Stephen Roller, Naman Goyal, Mikel Artetxe, Moya Chen, Shuohui Chen, Christopher Dewan, Mona Diab, Xian Li, Xi~Victoria Lin, Todor Mihaylov, Myle Ott, Sam Shleifer, Kurt Shuster, Daniel Simig, Punit~Singh Koura, Anjali Sridhar, Tianlu Wang, and Luke Zettlemoyer.
\newblock Opt: Open pre-trained transformer language models, 2022.

\bibitem[Zhou et~al.(2024)Zhou, Liu, Xu, Iyer, Sun, Mao, Ma, Efrat, Yu, Yu, et~al.]{zhou2024lima}
Chunting Zhou, Pengfei Liu, Puxin Xu, Srinivasan Iyer, Jiao Sun, Yuning Mao, Xuezhe Ma, Avia Efrat, Ping Yu, Lili Yu, et~al.
\newblock Lima: Less is more for alignment.
\newblock \emph{Advances in Neural Information Processing Systems}, 36, 2024.

\end{thebibliography}

\clearpage
\newpage

\appendix


\section{Codebase and Data}
We make all dataset splits and scripts to download the data and run baseline experiments open-source, available at \url{datas3-benchmark.github.io}.

\section{Additional Results}

\label{appdx:additional_results}
\subsection{Sample Efficiency}

\begin{figure*}[h]
    \centering
    \includegraphics[width=\linewidth]{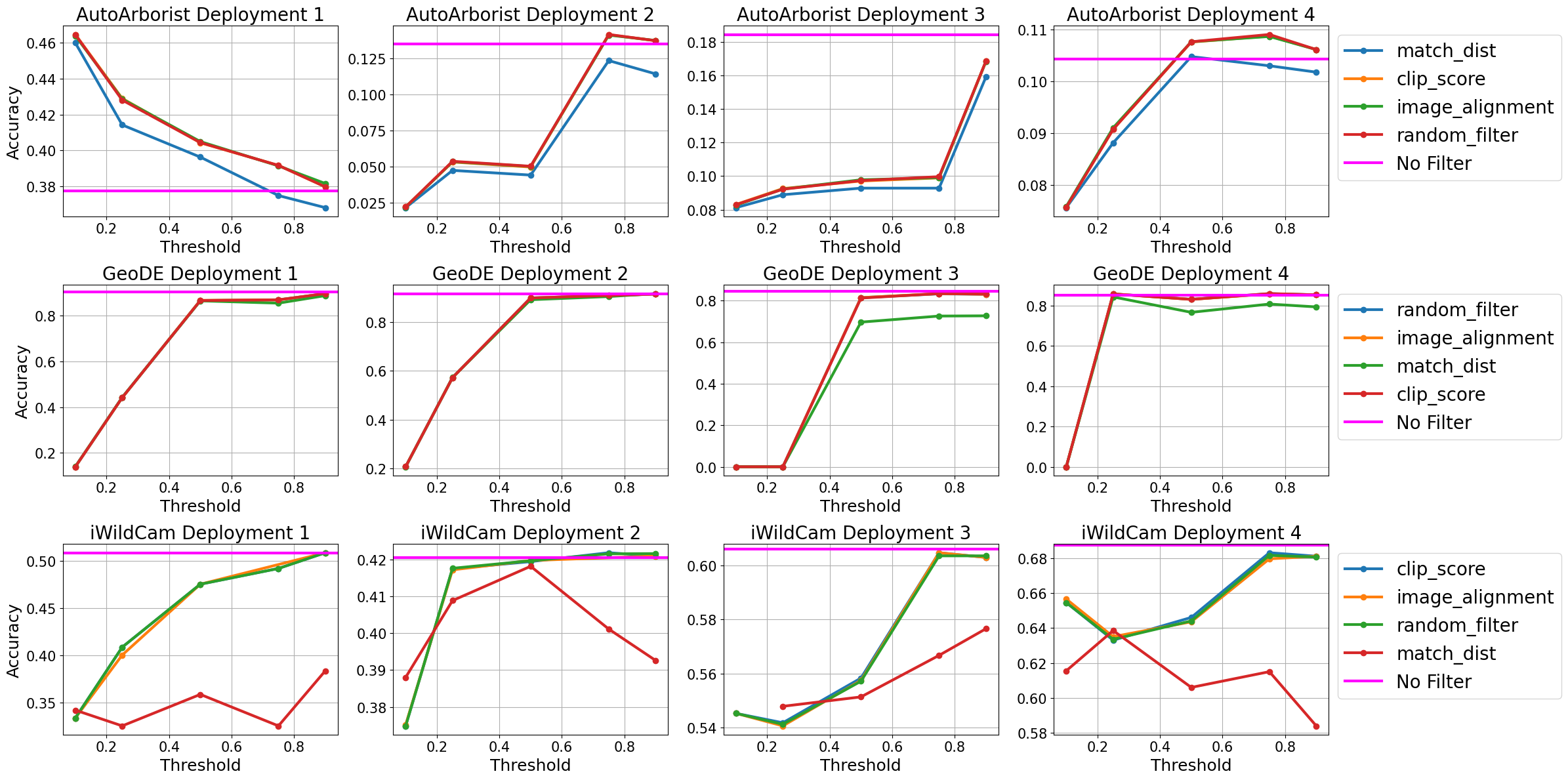}
    \caption{Plotting the sample efficiency of the baselines, for the baselines that thresholds of subset size were set (CLIP-score, Image-Align, Match-Dist), with Random as a comparison point) for the linear-probing results. NuScenes was left out because it uses MSE as a performance metric and cannot use the Match-Dist baseline because it is a regression baseline. We find that oftentimes, models perform nearly just as well with 50\% of the data, with examples of certain subsets outperforming using all the data.}
    \label{fig:efficiency_plot}
\end{figure*}

We plot the results of the baselines that had set thresholds of subset size in Figure \ref{fig:efficiency_plot} for the linear probing results. For 9 of the 16 deployments, there was no subset from the threshold baselines that outperformed using all the data. However, for the examples of AutoArborist Deployment 1, 2, and 4; GeoDE Deployment 3 and 4; and iWildCAM Deployment 1 and 2, there was a simple threshold baseline that outperformed all the data. In particular, AutoArborist Deployment 1 greatly benefitted from subsetting, likely due to its extremely long-tailed nature, as shown in Figure \ref{fig:autoarborist_deployments}. These "efficiency-style" experiments were not performed for the full-finetuning ResNet50 models, due to computational constraints of training a great number of models.

\subsection{Other Model Experiments}
\textbf{ViT Linear Probes:} In addition to full-finetuning of ResNets, we also performed linear probing of ViT embeddings as described in Section \ref{appdx:additional_training_details}. In contrast to full-finetuning, the gains made from well-chosen subsets to using all the data are minimal, with a maximum accuracy gain of $18\%$ for AutoArborist. In contrast to the full-finetuning results, there does not always exist a well-chosen subset for all deployments. We hypothesize that the model is not sufficiently updated for dataset subset selection when using linear probes - since the embeddings from the pretrained ViT model remain the same, the data does not play as much of an impact. Additionally, linear probes overall perform much worse than a ResNet full-finetune, with the notable exception of GeoDE. This is likely because GeoDE is an object recognition dataset, closer to the pretrained ImageNet-21k weights that the ViT model is well-posed to recognize. AutoArborist, iWildCam, and NuScenes are significantly OOD from the pretraining dataset of ViT and as such, linear probes perform poorly -- more training/finetuning is required for optimal performance. Additionally, the scale/size of the subset is much more important with linear probing than with full-finetuning -- this is perhaps why the query sets and expert subsets (which are quite small relative to the entire training dataset) perform poorly relative to training on all data. In contrast, a ResNet50 does not need to be trained on much data (such as the query/expert subsets) to perform near-optimal, if the data is nearly in-distribution.

\noindent\textbf{ViT LoRA Finetuning:} We also performed ViT LoRA finetuning as described in Section \ref{appdx:additional_training_details}. We see very similar results to the ResNet full-finetuning, largely with greater performance on some deployments than with ResNet (likely due to the larger model architecture).

\begin{table*}[]
\resizebox{\linewidth}{!}{
\begin{tabular}{@{}c|c|c|cc|c|cc|ccc@{}}
\specialrule{.2em}{.1em}{.1em} 
\multicolumn{1}{c|}{\multirow{2}{*}{Dataset}} & \multicolumn{1}{c|}{\multirow{2}{*}{Metric}} & \multicolumn{1}{l|}{\multirow{2}{*}{Deploy \#}} & \multicolumn{2}{c|}{Non subset} & 
\multicolumn{1}{c|}{\multirow{2}{*}{Knowledge-driven}} & \multicolumn{2}{c|}{Unlabeled query set} & 
\multicolumn{3}{c}{Labeled query set} 
\\ \cline{4-5} \cline{7-11}
\multicolumn{1}{l|}{} & 
\multicolumn{1}{l|}{} & 
\multicolumn{1}{l|}{} &
Query-set & All-data &  & Image-Align & Near-Nbors & CLIP-score & Match-Label & Match-Dist \\
\specialrule{.2em}{.1em}{.1em}
\multirow{4}{*}{GeoDE} & \multirow{4}{*}{Acc} &
   Deploy 1 & 0.851 & 0.904 & 0.083 & 0.869 & 0.867 & 0.897 & \textbf{0.904} & 0.887 \\
 & & Deploy 2 & 0.833 & 0.914 & 0.114 & 0.914 & 0.898 & 0.914 & 0.915 & \textbf{0.916 }\\
 & & Deploy 3 & 0.998 & 0.844 & 0.100 & 0.834 & 0.814 & 0.832 & \textbf{0.923} & 0.727 \\
 & & Deploy 4 & 0.969 & 0.853 & 0.123 & 0.858 & 0.837 & 0.854 & \textbf{0.900}& 0.794 \\
  \hline 
\multirow{4}{*}{iWildCam} & \multirow{4}{*}{Acc} &
   Deploy 1 & 0.858 & \textbf{0.509}& 0.350 & 0.508 & 0.391 & 0.508 & 0.350 & 0.383 \\
 & & Deploy 2 & 0.585 & 0.420 & 0.346 & 0.423 & 0.425 & 0.421 & \textbf{0.438} & 0.418 \\
 & & Deploy 3 & 0.824 & \textbf{0.606} & 0.550 & 0.604 & 0.569 & 0.604 & 0.554 & 0.567 \\
 & & Deploy 4 & 0.906 & \textbf{0.688} & 0.611 & 0.681 & 0.621 & 0.683 & 0.624 & 0.639 \\
  \hline 
\multirow{4}{*}{Auto Arborist} & \multirow{4}{*}{Acc} &
   Deploy 1 & 0.373 & 0.377 & \textbf{0.545} & 0.464 & 0.392 & 0.464 & 0.401 & 0.459 \\
 & & Deploy 2 & 0.109 & 0.134 & \textbf{0.245} & 0.137 & 0.141 & 0.137 & 0.139 & 0.123 \\
 & & Deploy 3 & 0.169 & 0.184 & \textbf{0.234} & 0.167 & 0.099 & 0.168 & 0.103 & 0.159 \\
 & & Deploy 4 & 0.125 & 0.104 & \textbf{0.300} & 0.106 & 0.108 & 0.108 & 0.108 & 0.108 \\
  \hline 
\multirow{4}{*}{NuScenes} & \multirow{4}{*}{MSE} &
   Deploy 1 & 0.494 & 0.508 & 0.730 & \textbf{0.333} & 0.391 & 0.475 & - & - \\
 & & Deploy 2 & 0.226 & 0.420 & 1.160 & \textbf{0.375} & 0.425 & 0.422 & - & - \\
 & & Deploy 3 & 0.559 & 0.606 & 1.112 & \textbf{0.540} & 0.566 & 0.558 & - & - \\
 & & Deploy 4 & 0.434 & 0.688 & 1.137 & 0.635 & \textbf{0.621} & 0.646 &- & - \\
\specialrule{.2em}{.1em}{.1em}
\end{tabular}
}
\caption{
Best-performing subsets across hyperparameters for baseline methods across all datasets and deployments (abbreviated as Deploy) for \textbf{ViT linear probes}. Overall accuracy is reported for the classification tasks of GeoDE, iWildCam, and Auto Arborist (greater is better) and MSE is reported for the regression task of NuScenes (smaller is better). Match-Dist and Match-Label are not applicable for NuScenes, as it is a regression task and does not have clear classes/labels for these methods. Baselines are distinguished from one another by their access to information, with each baseline having access to expert knowledge, or a labeled/unlabeled query set.}
\label{tab:linear_probe_results}
\end{table*}

\begin{table*}[]
\resizebox{\linewidth}{!}{
\begin{tabular}{@{}c|c|c|cc|c|cc|ccc@{}}
\specialrule{.2em}{.1em}{.1em} 
\multicolumn{1}{c|}{\multirow{2}{*}{Dataset}} & \multicolumn{1}{c|}{\multirow{2}{*}{Metric}} & \multicolumn{1}{l|}{\multirow{2}{*}{Deploy \#}} & \multicolumn{2}{c|}{Non subset} & 
\multicolumn{1}{c|}{\multirow{2}{*}{Knowledge-driven}} & \multicolumn{2}{c|}{Unlabeled query set} & 
\multicolumn{3}{c}{Labeled query set} 
\\ \cline{4-5} \cline{7-11}
\multicolumn{1}{l|}{} & 
\multicolumn{1}{l|}{} & 
\multicolumn{1}{l|}{} &
Query-set & All-data &  & Image-Align & Near-Nbors & CLIP-score & Match-Label & Match-Dist \\
\specialrule{.2em}{.1em}{.1em}
\multirow{4}{*}{GeoDE} & \multirow{4}{*}{Acc} & Deploy 1 & 0.870 & 0.871 & 0.878 & \textbf{0.895} & 0.883 & 0.890 & 0.882 & 0.886  \\
 & & Deploy 2 & 0.440 & 0.894 & 0.859 & 0.896 & 0.895 & \textbf{0.902} & 0.900 & 0.882  \\
 & & Deploy 3 & 0.961 & 0.758 & \textbf{0.994} & 0.837 & 0.799 & 0.894 & 0.83 & 0.879  \\
 & & Deploy 4 & 0.821 & 0.858 & \textbf{0.958} & 0.779 & 0.833 & 0.882 & 0.841 & 0.830  \\ \hline
\multirow{4}{*}{iWildCam} & \multirow{4}{*}{Acc} & Deploy 1 & 0.703 & 0.635 & 0.641 & 0.562 & 0.503 & 0.505 & 0.736 & \textbf{0.753}  \\
 & & Deploy 2 & 0.730 & 0.332 & 0.344 & 0.433 & 0.460 & 0.460 & 0.350 & \textbf{0.494}  \\
 & & Deploy 3 & 0.434 & 0.723 & 0.743 & 0.537 & 0.456 & 0.421 & 0.733 & \textbf{0.755}   \\
 & & Deploy 4 & 0.444 & 0.650 & 0.689 & 0.600 & 0.601 & 0.156 & 0.567 & \textbf{0.732}  \\ \hline
\multirow{4}{*}{AutoArborist} & \multirow{4}{*}{Acc}&  Deploy 1 & 0.154 & 0.450 & \textbf{0.888} & 0.381 & 0.390 & 0.388 & 0.665 & 0.777   \\
 & & Deploy 2 & 0.233 & 0.489 & \textbf{0.945} & 0.133 & 0.141 & 0.138 & 0.645 & 0.564   \\
 & & Deploy 3 & 0.138 & 0.169 & \textbf{0.384} & 0.179 & 0.129 & 0.169 & 0.156 & 0.234   \\
 & & Deploy 4 & 0.147 & 0.155 & \textbf{0.452} & 0.104 & 0.118 & 0.146 & 0.152 & 0.233   \\ \hline
\multirow{4}{*}{NuScenes} & \multirow{4}{*}{MSE} & Deploy 1 & 0.061 & 0.034 & \textbf{0.034} & 0.044 & 0.055 & 0.072 & - & -   \\
 & & Deploy 2 & 0.071 & \textbf{0.011} & 0.045 & 0.167 & 0.056 & 0.033 & - & -   \\
 & & Deploy 3 & 0.099 & 0.045 & \textbf{0.034} & 0.041 & 0.122 & 0.072 & - & -   \\
 & & Deploy 4 & 0.120 & 0.049 & \textbf{0.040} & 0.084 & 0.399 & 0.050 & - & -   \\ 
 \specialrule{.2em}{.1em}{.1em}
\end{tabular}
}
\caption{Best-performing subsets across hyperparameters for baseline methods across all datasets and deployments (abbreviated as Deploy) for \textbf{ViT LoRA finetuning}. Overall accuracy is reported for the classification tasks of GeoDE, iWildCam, and Auto Arborist (greater is better) and MSE is reported for the regression task of NuScenes (smaller is better). Match-Dist and Match-Label are not applicable for NuScenes, as it is a regression task and does not have clear classes/labels for these methods. Baselines are distinguished from one another by their access to information, with each baseline having access to expert knowledge, or a labeled/unlabeled query set.}
\label{tab:lora_results}
\end{table*}

\section{Additional Training Information}
\label{appdx:additional_training_details}
\noindent\textbf{(1) Subset selection step:} Users must select data from the general training pool for a deployment with the given query set. Query sets are split from the evaluation/deployment set with a stratified sampling strategy, ensuring that at least two of each class is in the query set. This is done to ensure that long-tailed datasets (such as iWildCam and AutoArborist) have all classes present in the query set to be representative of the deployment. iWildCam, NuScenes, and GeoDE have 500 sample query sets, whereas AutoArborist has a 1500 sample query set (because of its extremely long-tailed nature and the complexity of the dataset). 

\noindent\textbf{(2) Training/Fine-tuning step:} For all classification datasets, we use ResNet50 \cite{He2015DeepRL} for full-finetuning, a ViT \cite{Dosovitskiy2020AnII} for LoRA finetuning and for linear probes of the training subsets, chosen for efficiency due to the number of baselines. No additional data augmentation was performed to fully understand the role of the subset quality by itself. For full-finetuning, we use the Adam optimizer with cross-entropy loss, starting from ImageNet-1k weights and updating all layers. For the linear probing, we train a \texttt{sk-learn} logistic/linear regression classifier on top of ViT-B/32 features pretrained from ImageNet-21k and labels for classification/regression tasks, respectively. For LoRA finetuning, we use the \texttt{peft} package to select layers to finetune/ We run a small hyperparameter sweep for each baseline across batch sizes $\{32,64,128\}$ and learning rates $\{0.01,0.001,0.0001\}$ for each deployment across a validation set, split from the training subset in a random $90/10$ split. All models were trained on A100 GPU's. All models were implemented using PyTorch.

For detection tasks, we use YOLOv8n \cite{Jocher_Ultralytics_YOLO_2023} model with default training parameters, such as learning rates/schedules and default COCO-pretrained models, for 100 epochs, and subsample images to 640p resolutions, chosen for efficiency due to computational costs of training detectors. These models were trained across 4 L40 GPU's in parallel.

\section{Additional Dataset Details}
\label{appdx:additional_dataset_details}
\subsection{iWildCam}
\noindent\textbf{Additional dataset information.} These images tend to be taken in short bursts following the motion-activation of a camera trap, so the images can be additionally grouped into sequences of images from the same burst, though our baseline models do not exploit this information, and our evaluation metric treats each image individually. However, a grouped sequence is in the same split of the data (train, test, query) in order to avoid model memorization. Each image is associated with the following metadata: camera trap ID, sequence ID, and datetime. 

\begin{figure*}[h]
    \centering
    \includegraphics[width=\linewidth]{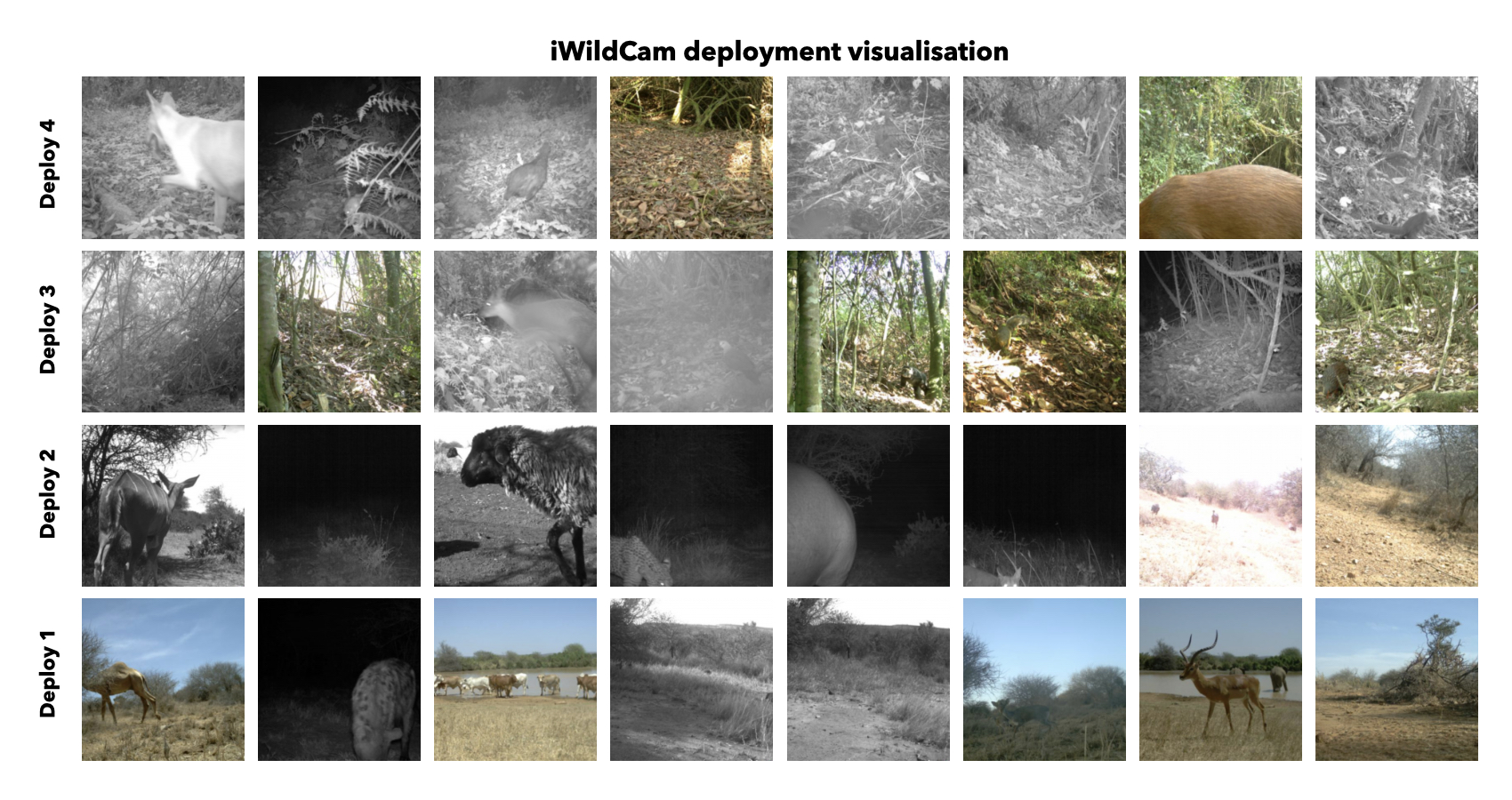}
    \caption{Visualization of the iWildCam dataset across deployments}
    \label{fig:iWildCam_viz}
\end{figure*}

\begin{figure*}
    \centering
    \includegraphics[width=0.75\linewidth]{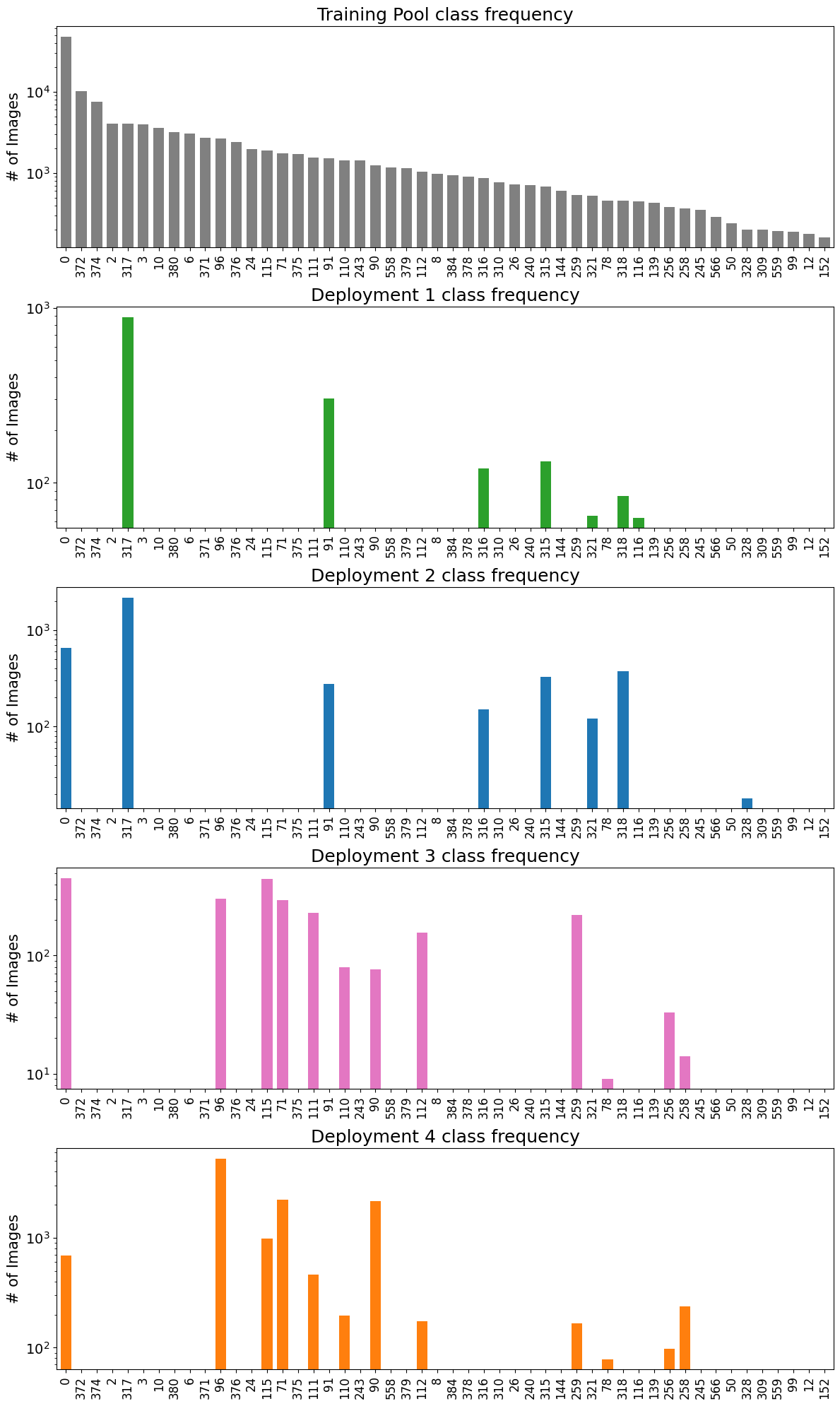}
    \caption{iWildCam deployment label distribution for the 50 most common classes (common determined by the training pool) in log-scale. As seen, there is significant label shift from the training pool to the deployments, and from the deployments to each other.}
    \label{fig:iwildcam_deployments}
\end{figure*}

\noindent\textbf{Deployment splits. } The splits for each deployment were created by mapping the locations of the iWildCam camera traps to latitude and longitude, then clustering geospatially, where every cluster formed a 100km radius. The deployment splits are subsets of the testing split of the WILDS-iWildCam~\cite{Koh2020WILDSAB} dataset. In~ \cite{Koh2020WILDSAB}, the "in-distribution" test split (\texttt{test-id}) was split from the training set by the time of the capture, whereas the "out-of-distribution" test split (\texttt{test-ood}) where was split from the training set by location. Deployments 1 and 3 are from the original \texttt{test-id} split of the dataset and Deployments 2 and 4 are from the \texttt{test-ood} split. As a result, Deployments 1 and 3 are ID dataset subselection tasks -- meaning that data from the locations in these deployments exists in the training pool. In contrast, Deployments 2 and 4 are OOD dataset subselection tasks -- there is no existing data from these locations in the training pool. We expect that OOD tasks will be more difficult to find well-performing subsets for. Additional details about each deployment are given in Table \ref{tab:iwildcam_deployments} and visualized in Figure \ref{fig:iwildcam_deployments} and Figure \ref{fig:iWildCam_viz}.

\noindent\textbf{Expert subsets.  } For Deployments 1 and 3 (the ID deployments), the expert subsets were created by only choosing data from the relevant locations in the testing pool (these locations are given in Table \ref{tab:iwildcam_deployments}) and eliminating irrelevant classes that are not present in the testing set. For Deployments 2 and 4 (the OOD deployments), the expert subsets were created by only choosing data from locations that are within 500km of the locations in the deployments and eliminating irrelevant classes that are not present in the testing set.

\begin{table*}[] 
\begin{tabular}{lllp{3cm}p{6cm}}
\hline
Deployment \# & \# classes & \# images & locations & class\_label\\
\hline 
Deployment 1  & 14 & 1744 & 225, 333, 312, 530, 416, 541, 25, 356, 504, 521, 71, 202, 516, 224, 162 & 
funisciurus carruthersi, cricetomys gambianus, cephalophus nigrifrons, protoxerus stangeri, atherurus africanus, turtur tympanistria, francolinus nobilis, potamochoerus larvatus, cercopithecus lhoesti, pan troglodytes, cercopithecus mitis, francolinus africanus, hylomyscus stella, canis lupus\\ \hline
Deployment 2  & 15 & 4348 & 167, 282, 385, 123, 110, 144, 231, 417, 314 
& francolinus nobilis, cephalophus nigrifrons, empty, funisciurus carruthersi, atherurus africanus, cricetomys gambianus, cercopithecus lhoesti, genetta tigrina, cercopithecus mitis, paraxerus boehmi, pan troglodytes, genetta servalina, mus minutoides, cephalophus silvicultor, turtur tympanistria\\ \hline
Deployment 3  & 30 & 2558 & 242, 51, 101, 508, 372, 522, 247, 138, 359, 379, 91, 157, 410, 259, 369, 73, 220, 415, 518, 387, 10, 170,  226, 443, 383, 251, 450, 532, 503, 81, 437, 151, 127, 113, 471, 197, 112, 130, 65, 273, 254, 366, 400, 31, 551, 298, 24, 124, 169, 318, 159 
& empty, aepyceros melampus, madoqua guentheri, ichneumia albicauda, kobus ellipsiprymnus, syncerus caffer, leptailurus serval, bos taurus, giraffa camelopardalis, loxodonta africana, equus grevyi, crocuta crocuta, equus quagga, capra aegagrus, felis silvestris, panthera pardus, camelus dromedarius, proteles cristata, phacochoerus africanus, acryllium vulturinum, papio anubis, ovis aries, hippopotamus amphibius, tragelaphus scriptus, unknown bird, eupodotis senegalensis, tragelaphus strepsiceros, lepus saxatilis, tragelaphus oryx, lissotis melanogaster\\ \hline
Deployment 4  & 33 & 13931 
& 501, 306, 206, 499, 131, 53, 140, 375, 179, 193, 291, 229, 249, 108, 477, 328, 152, 302, 278, 329, 422, 540, 520, 136, 489, 492
& bos taurus, loxodonta africana, equus quagga, aepyceros melampus, empty, phacochoerus africanus, oryx beisa, tragelaphus oryx, lycaon pictus, syncerus caffer, crocuta crocuta, madoqua guentheri, acryllium vulturinum, capra aegagrus, caracal caracal, lepus saxatilis, hippopotamus amphibius, giraffa camelopardalis, papio anubis, camelus dromedarius, hystrix cristata, equus grevyi, panthera leo, ovis aries, panthera pardus, proteles cristata, hyaena hyaena, canis lupus, canis mesomelas, leptailurus serval, kobus ellipsiprymnus, eupodotis senegalensis, lissotis melanogaster\\
\hline
\end{tabular}
\caption{iWildCam Deployment split details. As seen, the deployments have severe label shift from one another and come from disjoint locations. They also vary in size and the number of classes.}
\label{tab:iwildcam_deployments}
\end{table*}

\newpage
\subsection{GeoDE}
\noindent\textbf{Additional dataset information.} Datasets used for object classification tasks are often constructed by scraping images from the web. Examples of such datasets include ImageNet \cite{imagenet,Russakovsky2014ImageNetLS}, Open Images \cite{Kuznetsova2018TheOI}, PASS \cite{Asano2020NewCD}, CLIP-400M \cite{Brown2020LanguageMA} etc. Constructing such datasets is cheap,  and thus scalable, however, such datasets are known to contain, among others, geographical biases \cite{Shankar2017NoCW}. Rather than scraping images from the web, GeoDE \cite{ramaswamy2023geode} crowdsources a dataset roughly balanced across 40 different objects and 6 world regions. Crowdsourcing a dataset allows for tighter control over the data distribution. For example, it allows us to target specific regions and objects that are underrepresented within webscraped datasets. However, it can also be prohibitively expensive, limiting the size of such datasets. Thus, it becomes paramount to understand which objects and regions should be targeted within crowdsourced data collection. 

\begin{figure*}[h]
    \centering
    \includegraphics[width=\linewidth]{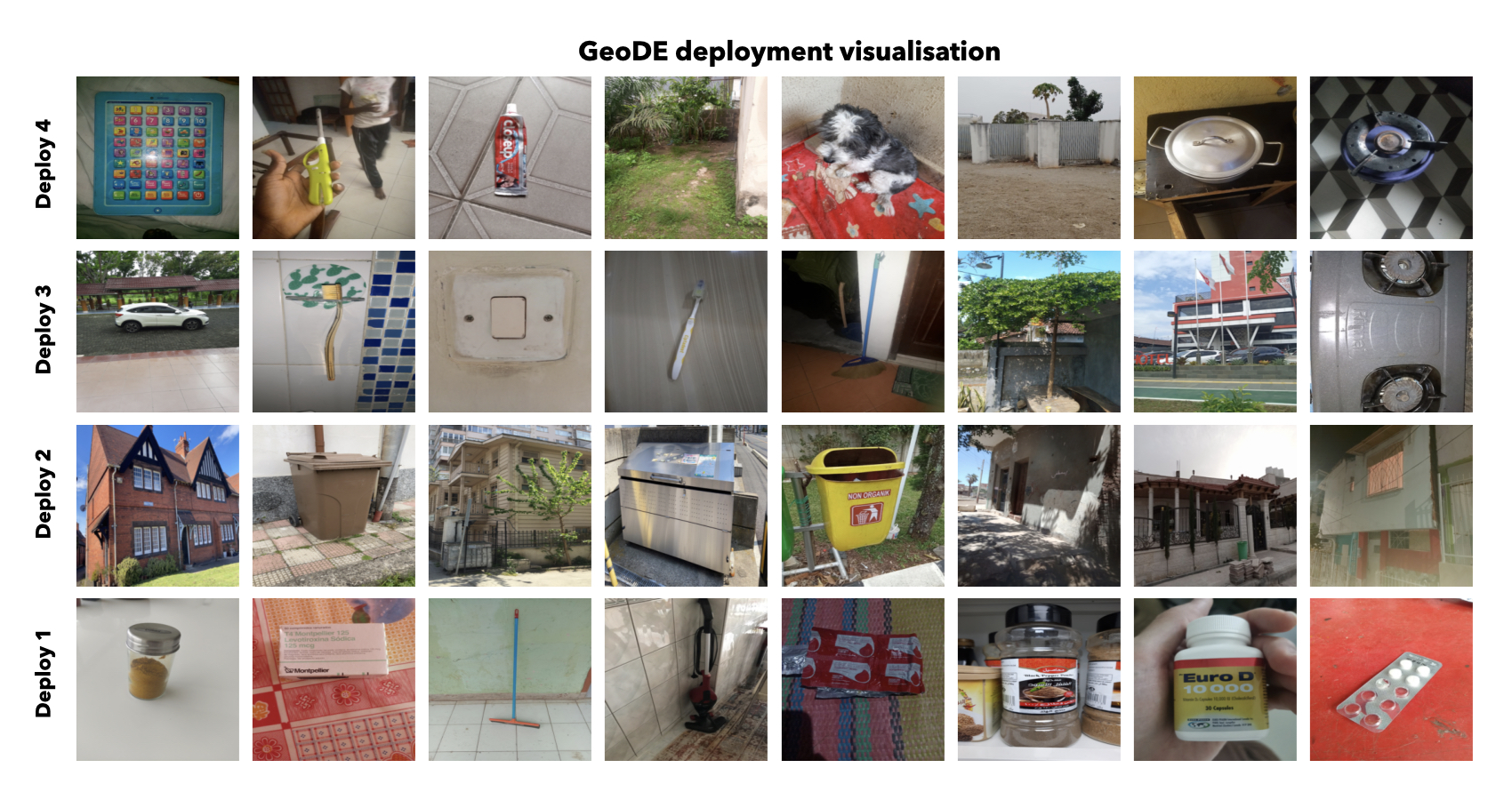}
    \caption{Visualization of the GeoDE dataset across deployments}
    \label{fig:GeoDE_viz}
\end{figure*}

\noindent\textbf{Deployment splits.} There are two types of deployments: country and object splits. As mentioned, Deployments 1 and 2 are Nigeria and Indonesia, respectively, chosen because they are the two countries with the regions with the poorest performance in the original GeoDE paper \cite{ramaswamy2023geode}. Similarly, Deployments 3 and 4 are of indoor/outdoor objects, respectively, with the worst performing classes chosen for this deployment. In the benchmark, we also include an additional potential pool of training data not explored in this experimental study -- the "low data quality" pool not included in the original GeoDE paper. Users of this benchmark are allowed to add this data to their subset. The deployment's images are visualized in Figure \ref{fig:GeoDE_viz}.

\noindent\textbf{Expert subsets.  } For Deployments 1 and 3 (the ID deployments), the expert subsets were created by only choosing data from the relevant locations in the testing pool (these locations are given in Table \ref{tab:iwildcam_deployments}) and eliminating irrelevant classes that are not present in the testing set. For Deployments 2 and 4 (the OOD deployments), the expert subsets were created by only choosing data from locations that are within 500km of the locations in the deployments and eliminating irrelevant classes that are not present in the testing set.

\begin{table*}[] 
\begin{tabular}{lllp{10cm}}
\hline
Deployment \# & \# classes & \# images  & class\_label\\
\hline 
(1) Nigeria  & 40 & 2155 & 
bag, hand soap, dustbin, toothbrush, toothpaste toothpowder, hairbrush comb, chair, hat, light fixture, light switch, plate of food, spices, stove, cooking pot, cleaning equipment, lighter, medicine, candle, toy, jug, streetlight lantern, front door, tree, house, backyard, truck, waste container, car, fence, road sign, dog, wheelbarrow, religious building, stall, boat, monument, flag, bus, storefront, bicycle\\ 
\hline
(2) Indonesia  & 40 & 4348 & bag, hand soap, dustbin, toothbrush, toothpaste toothpowder, hairbrush comb, chair, hat, light fixture, light switch, plate of food, spices, stove, cooking pot, cleaning equipment, lighter, medicine, candle, toy, jug, streetlight lantern, front door, tree, house, backyard, truck, waste container, car, fence, road sign, dog, wheelbarrow, religious building, stall, boat, monument, flag, bus, storefront, bicycle
\\ \hline
(3) Indoor Objects  & 2 & 924 & house, waste container\\ \hline
(4) Outdoor Objects  & 3 & 2116
& spices, cleaning equipment, medicine\\
\hline
\end{tabular}
\caption{GeoDE Deployment split details. As seen, the deployments have label shift from one another, varying in size and class labels.}
\label{tab:geode_deployments}
\end{table*}

\begin{figure*}
    \centering
    \includegraphics[width=0.75\linewidth]{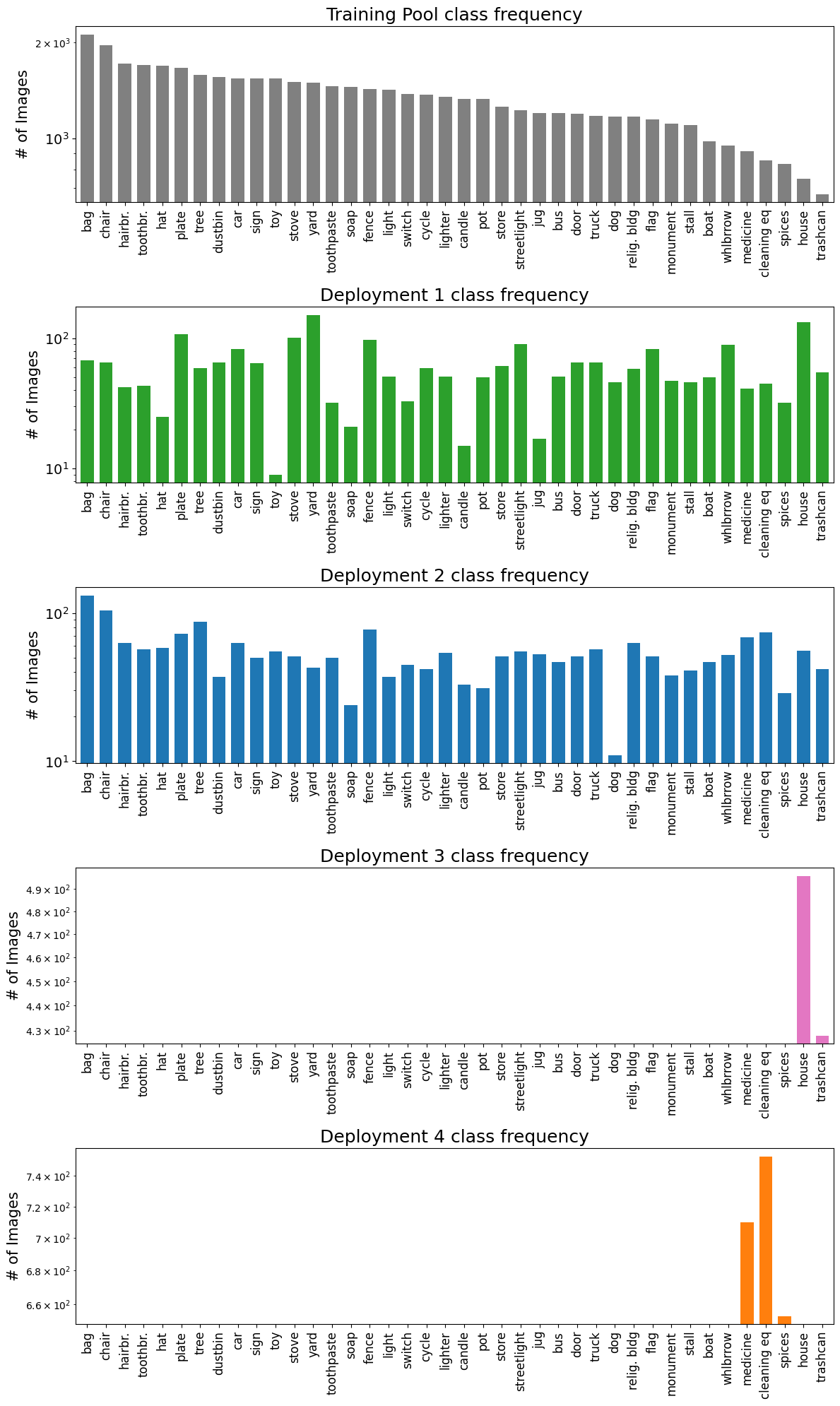}
    \caption{GeoDE deployment label distribution for all classes. As seen, there is significant label shift from the training pool to the deployments, and from the deployments to each other.}
    \label{fig:geode_deployments}
\end{figure*}

\newpage
\subsection{AutoArborist}
\noindent\textbf{Additional dataset information.} Environmental monitoring and Earth observation from aerial imagery have the potential to enable policymakers to make data-informed decisions to facilitate societal adaptation to a changing climate \cite{Brandt2016AFF}. However, aerial/street-level data repositories from satellite and low-flying aircraft are currently in the petabyte scale and growing, making extracting useful and relevant information to support policy intractable without automation. Tree image classification has potential impact on humanitarian aid and disaster relief, wilderness forests, agriculture, and urban mapping with uses in city planning, resource management, and environmental monitoring. For example, urban ecologists need to know the location and type of trees in cities so that they can target replanting to improve climate adaptation. Collecting this information from ground-level tree censuses is both time-consuming and expensive, thus automated tree genus classification from Global Positioning System (GPS)-registered aerial imagery is increasingly of interest. Datasets like AutoArborist enable the computer vision community to investigate automated methods for tree genus classification from aerial imagery at scale, containing images and genus labels for over 1M individual trees \cite{beery2022auto}.

Additional challenges in the data include:
\begin{itemize}
    \item \textbf{Noisy labels.} Images are commonly mislabeled: with genus classification is difficult and requires specialized expertise, GPS localization from the ground can be in error, there are often multiple trees within a single image with only a single label, and temporal inconsistencies can occur as trees are not imaged and labeled at the same time.
    \item \textbf{Non-IID data.} Geospatial data also breaks the typical deep learning assumption that data will be independent and identically distributed (IID) spatially close examples often contain correlations. For example, trees are often planted in groups (e.g. a row of cherry trees along the same street).
    \item \textbf{Fine-grained and long-tailed class distribution.} Tree classification is fine-grained, with only subtle differences between many genera, and the distribution of trees is long-tailed. These characteristics tend to skew classification models towards predicting predominant classes.
    \item \textbf{Geospatial distribution shift.} Finally, this dataset contains significant covariate and subpopulation distribution shift due to variations in weather, differences in urban planning specific to each city, and temporal changes at different locations.
\end{itemize}

\noindent\textbf{Deployment splits.} The splits for each deployment were created by city, where the evaluation splits are subsets of the testing split of the original AutoArborist~\cite{beery2022auto} dataset. Deployments 3 and 4 (Los Angeles and Washington DC) are the "in-distribution" (ID) deployments,  where the training data from these cities was left in the training pool. Deployments 1 and 2 (Surrey and Calgary) were the "out-of-distribution" (OOD) deployments, where there was no data from these cities in the training pool. The training pool additionally contains data from 19 other major cities in North America. 

\begin{figure*}
    \centering
    \includegraphics[width=\linewidth]{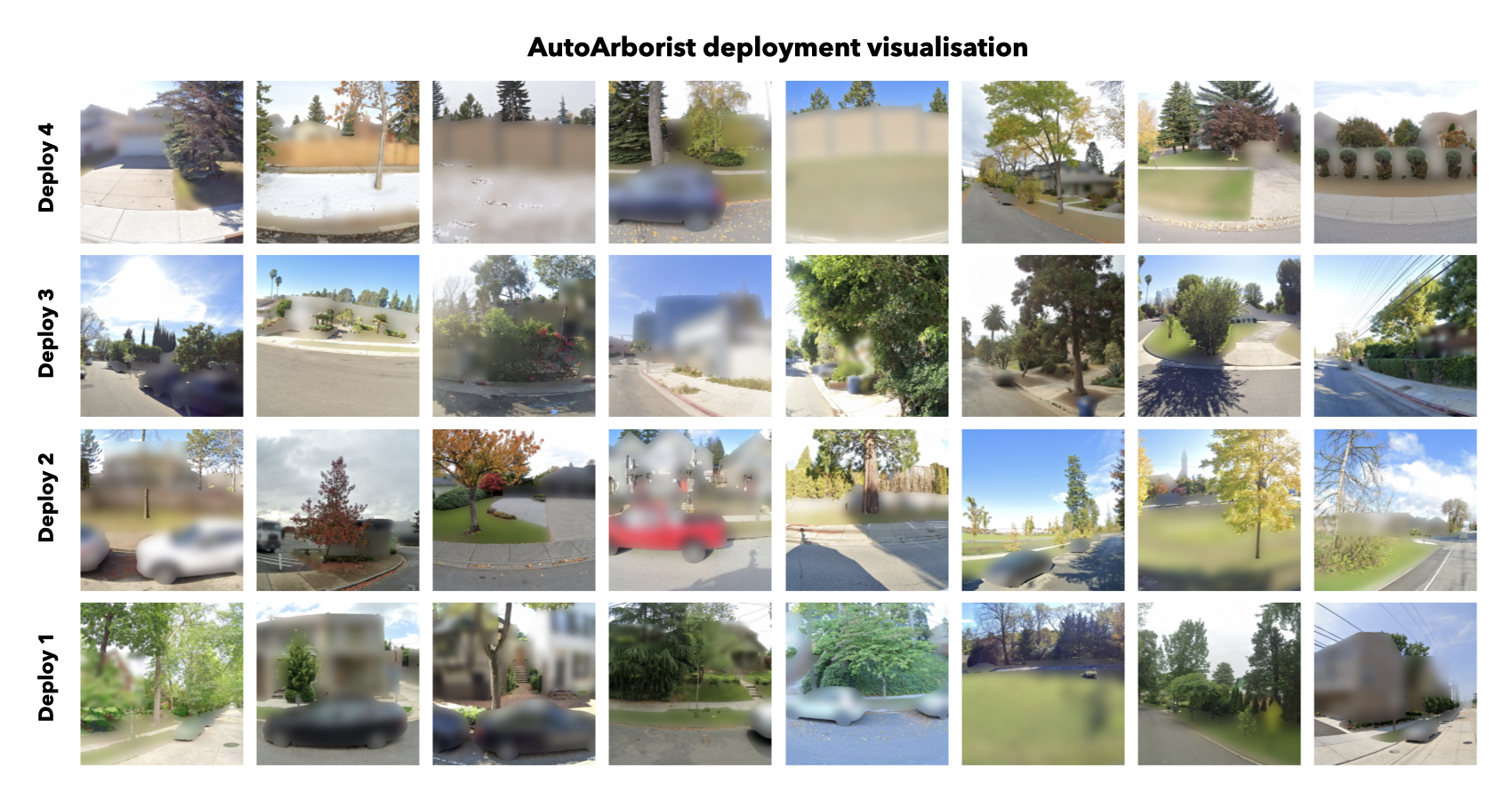}
    \caption{Visualization of the AutoArborist dataset across deployments}
    \label{fig:autoarborist_viz}
\end{figure*}

\begin{figure*}
    \centering
    \includegraphics[width=0.75\linewidth]{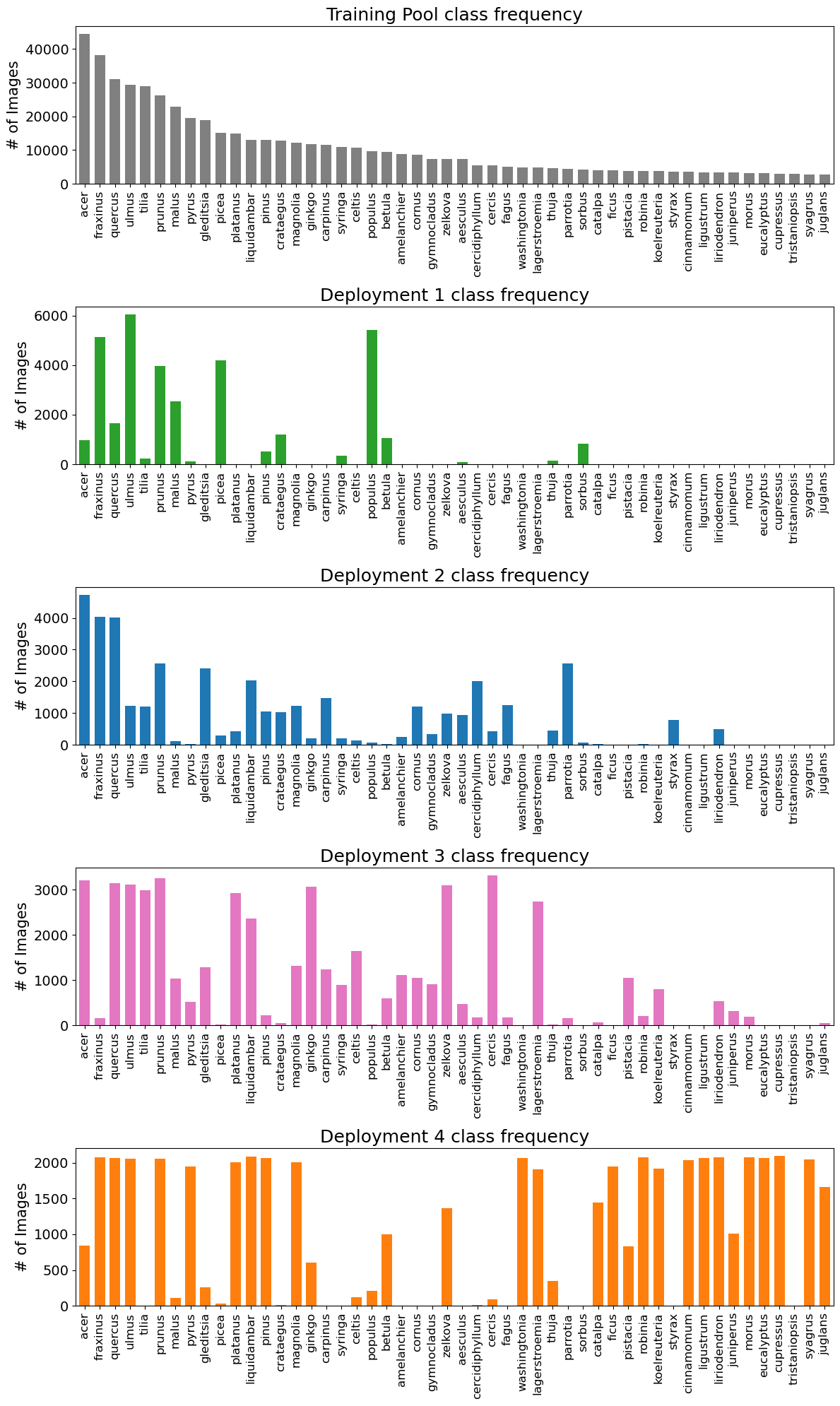}
    \caption{AutoArborist deployment label distribution for the 50 most common classes (commonly determined by the training pool) in log-scale. As seen, there is a significant label shift from the training pool to the deployments, and from the deployments to each other.}
    \label{fig:autoarborist_deployments}
\end{figure*}

\noindent\textbf{Expert subsets.} For expert subsets, we used data from the city itself for the ID deployments, and data from the closest cities for the OOD deployments. Accordingly, we used data from San Francisco and San Jose for Los Angeles and Charlottesville, Pittsburgh, and New York for Washington DC. We expect that OOD tasks will be more difficult to find well-performing subsets for in the benchmark. Label distribution shift is visualized in Figure \ref{fig:autoarborist_deployments} and covariate shift visualized in Figure \ref{fig:autoarborist_viz}.

\begin{table*}[] 
\begin{tabular}{lllp{9.5cm}}
\hline
Deployment & \# classes & \# images &  class\_label\\
\hline \hline
(1) Surrey  & 66 & 44295 & 
cladrastis, pinus, styrax, cornus, prunus, quercus, crataegus, liquidambar, ulmus, gleditsia, fraxinus, tilia, parrotia, platanus, stewartia, acer, aesculus, fagus, metasequoia, carpinus, zelkova, celtis, magnolia, liriodendron, amelanchier, cercidiphyllum, davidia, nyssa, syringa, pseudotsuga, thuja, cercis, phellodendron, chamaecyparis, ginkgo, calocedrus, gymnocladus, sorbus, sequoiadendron, picea, robinia, taxodium, catalpa, abies, malus, tsuga, eucommia, alnus, koelreuteria, betula, pyrus, populus, cedrus, sequoia, salix, juglans, pistacia, hibiscus, halesia, corylus, nothofagus, ilex, thujopsis, pterocarya, larix, paulownia\\ \hline

(2) Calgary  & 15 & 36347 & 
acer, ulmus, syringa, crataegus, sorbus, fraxinus, picea, prunus, populus, quercus, betula, malus, thuja, larix, tilia, pinus, elaeagnus, salix, pyrus, pseudotsuga, aesculus, abies, hippophae, caragana, juniperus, ribes, cotoneaster, alnus, tsuga, gleditsia\\ \hline

(3) Los Angeles  & 30 & 146849 & 
platanus, cedrus, ailanthus, nerium, syagrus, juglans, pinus, cupaniopsis, washingtonia, schinus, cinnamomum, lagerstroemia, jacaranda, libocedrus, syzygium, ceratonia, podocarpus, morus, liquidambar, zelkova, yucca, melaleuca, acacia, laurus, pistacia, fraxinus, robinia, magnolia, thuja, tabebuia, ligustrum, catalpa, koelreuteria, ulmus, salix, pittosporum, quercus, trachycarpus, casuarina, betula, rhus, callistemon, calocedrus, prunus, olea, archontophoenix, tristania, cassia, eucalyptus, citrus, lagunaria, ficus, phoenix, liriodendron, cordyline, malus, pyrus, celtis, cupressus, brachychiton, alnus, acer, ginkgo, juniperus, hymenosporum, photinia, eriobotrya, hibiscus, bauhinia, melia, thevetia, geijera, sequoia, sequoiadendron, maytenus, grevillea, erythrina, broussonetia, carya, tipuana, cercis, chionanthus, calodendrum, ceiba, chamaerops, sapium, diospyros, albizia, gleditsia, sambucus, musa, araucaria, strelitzia, vitex, psidium, cocos, dodonaea, metrosideros, heteromeles, populus, macadamia, sphaeropteris, eugenia, leptospermum, feijoa, platycladus, persea, casimiroa, sophora, dracaena, xylosma, livistona, schefflera, crinodendron, brahea, leucaena, ilex, arbutus, taxodium, punica, tamarix, butia, agonis, harpephyllum, nicotiana, rhaphiolepis, crataegus, plumeria, cycas, cornus, euonymus, lycianthes, myoporum, parkinsonia, picea, cercidiphyllum, elaeagnus, euphorbia, viburnum, quillaja, cotoneaster, pyracantha, paulownia, cocculus, caesalpinia, camellia, stenocarpus, lyonothamnus, maclura, osmanthus, beaucarnea, firmiana, castanea, umbellularia, wisteria, sorbus, metasequoia, myrtus, ziziphus, hakea, spathodea, annona, cryptomeria, olmediella, solanum, abies, aesculus, howea, ensete, carica, pseudotsuga, fremontodendron, chiranthodendron, chamaecyparis, cotinus\\ \hline

(4) Washington DC  & 33 & 71519 &
acer, tilia, syringa, celtis, gleditsia, ginkgo, cercis, amelanchier, platanus, malus, zelkova, prunus, nyssa, liquidambar, magnolia, lagerstroemia, pinus, ulmus, gymnocladus, quercus, carpinus, cladrastis, pyrus, betula, robinia, juniperus, koelreuteria, ilex, metasequoia, liriodendron, taxodium, cornus, styphnolobium, pistacia, morus, fagus, cedrus, aesculus, populus, crataegus, chionanthus, fraxinus, parrotia, laburnum, cercidiphyllum, rhus, catalpa, eucommia, halesia, ostrya, stewartia, sassafras, picea, cryptomeria, thuja, juglans, diospyros, cotinus, ailanthus, carya, oxydendrum, tsuga, salix, maclura, phellodendron, maackia, paulowniar\\
\hline
\end{tabular}
\caption{AutoArborist Deployment split details. As seen, the deployments have severe label shift from one another and come from disjoint locations. They also vary in size and the number of classes.}
\end{table*}

\newpage
\subsection{NuScenes}
\textbf{Additional data information.} The images were captured from a video stream recorded while driving a car. Each image is paired with a steering angle control from the CAN bus, synchronized with the sensor timestamps of both the camera and CAN bus data. To label each image with the correct steering angle, we apply 1D interpolation to create a continuous function of the steering angle and query it based on the camera’s timestamp. The steering angle, measured in radians, ranges from -7.7 to 6.3, with 0 indicating straight driving, positive values indicating left turns, and negative values indicating right turns. To ensure alignment between images and steering control data, samples with vehicle velocities below 1 m/s are removed.

\begin{figure*}
    \centering
    \includegraphics[width=\linewidth]{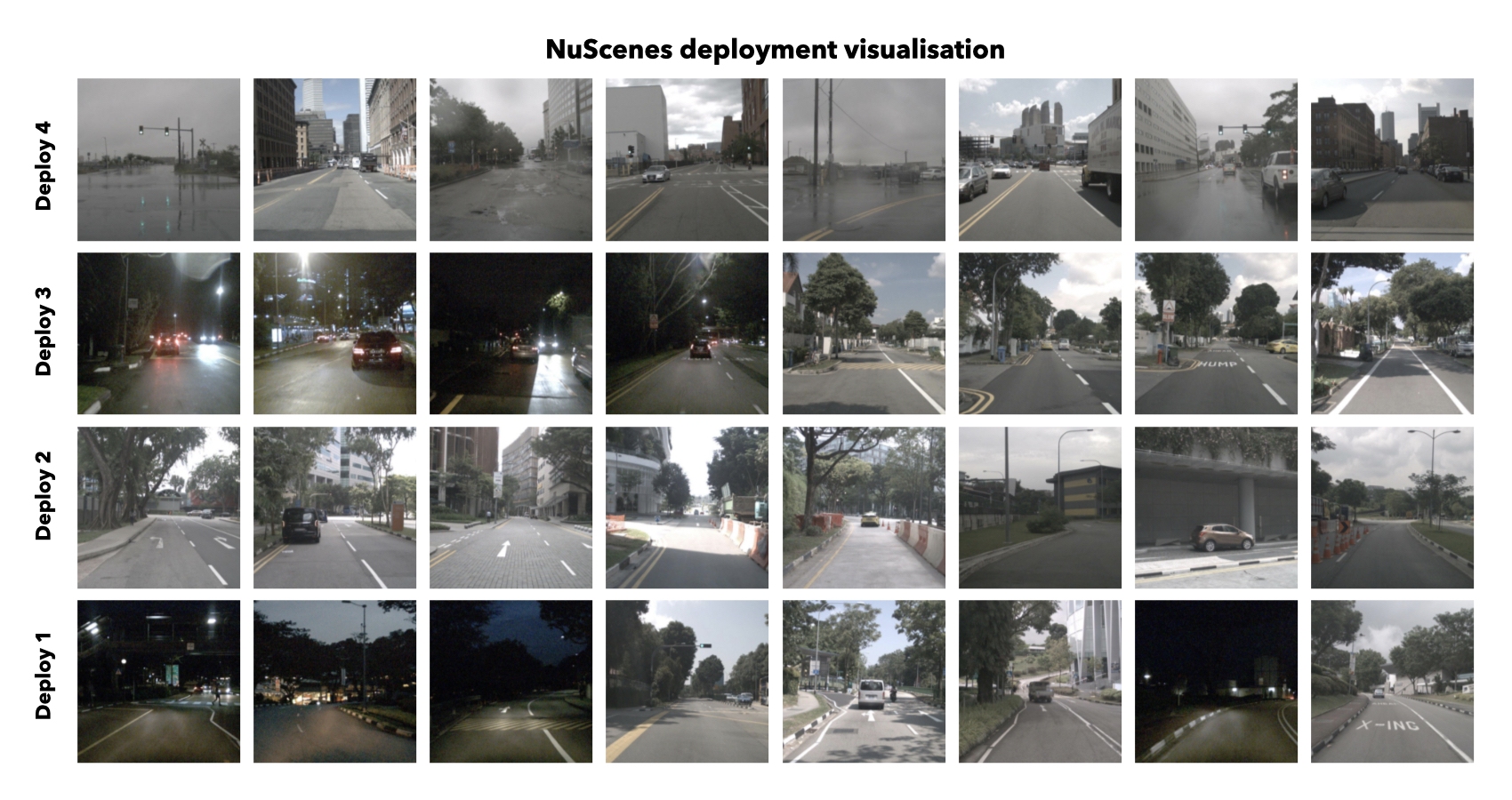}
    \caption{Visualization of the NuScenes dataset across deployments}
    \label{fig:nuscenes_viz}
\end{figure*}

\textbf{Deployment splits.} The splits for each deployment were created by city, where the evaluation splits are subsets of the testing split of the original NuScenes~\cite{caesar2020nuscenes} dataset. All deployments are in-distribution, meaning that there exists data from each city's deployment in the training pool. In contrast to AutoArborist, iWildCam, and GeoDE, where there is "extraneous" data in the training pool (that isn't necessarily relevant to the deployment geospatially or label-wise), there only exists data from the deployment cities in the training pool. The covariate shift in this dataset is visualized in Figure \ref{fig:nuscenes_viz}.

\begin{figure*}
    \centering
    \includegraphics[width=\linewidth]{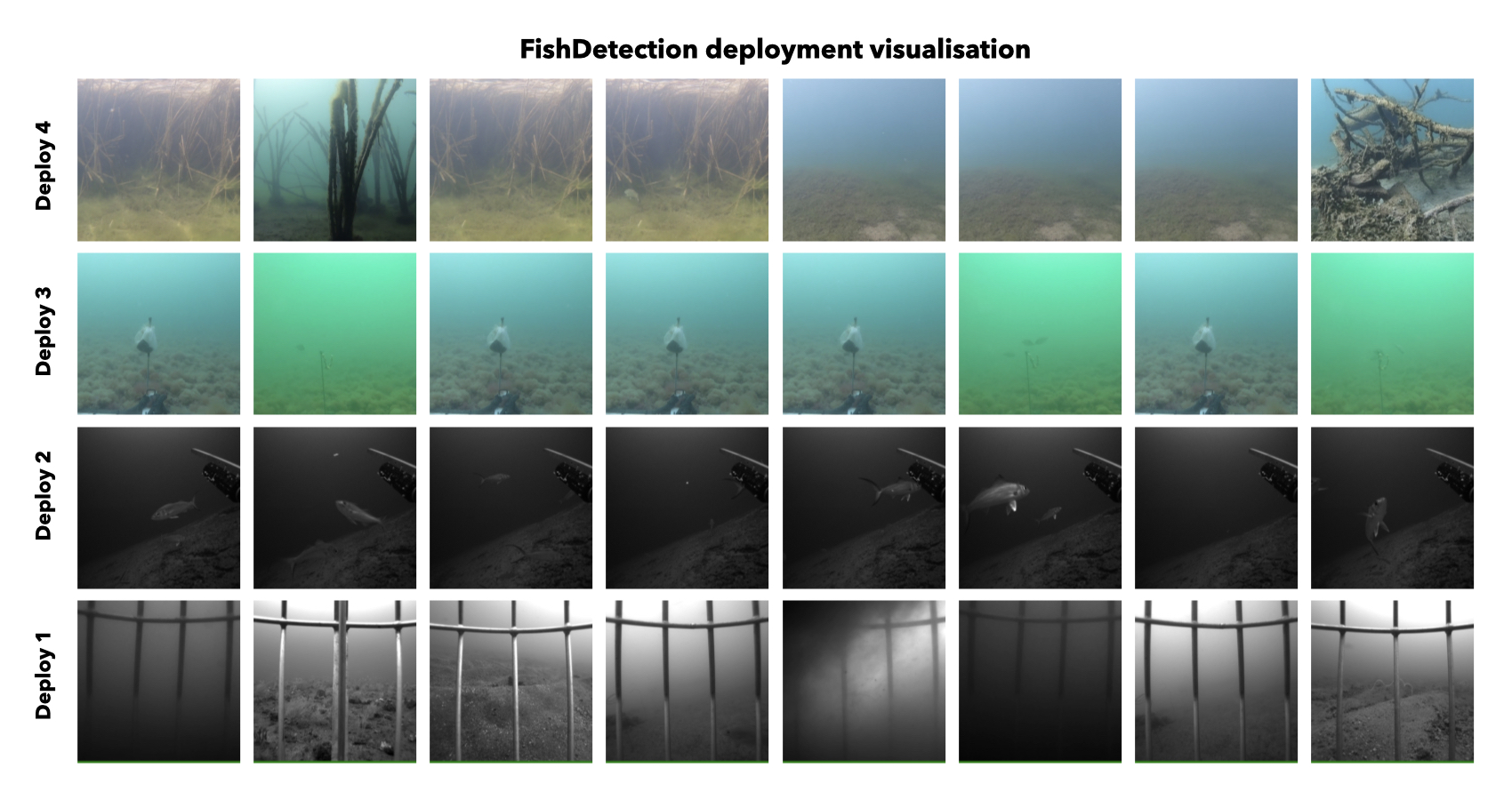}
    \caption{Visualization of the FishDetection dataset across deployments}
    \label{fig:fishdetection_viz}
\end{figure*}

\end{document}